\PassOptionsToPackage{table}{xcolor}

\documentclass[]{mpai_report}

\usepackage[toc,page,header]{appendix}

\usepackage{natbib}

\usepackage{CJKutf8}

\usepackage{xargs}  

\usepackage{todonotes}  

\usepackage{multirow}

\usepackage{cleveref}

\usepackage{amsmath}
\usepackage{dsfont}

\usepackage{subcaption}

\usepackage{svg}

\usepackage{mathrsfs}
\usepackage{adjustbox}
\usepackage{multirow}
\usepackage{multirow}
\usepackage{multicol}
\usepackage{tcolorbox}
\usepackage{changepage}
\usepackage{graphicx}
\usepackage{amssymb}
\usepackage{array}
\usepackage{bm}
\usepackage{hyperref}

\usepackage{minitoc}
\usepackage{pifont}
\newcommand{\cmark}{\textcolor[rgb]{0,0.6,0}{\ding{51}}}
\newcommand{\xmark}{\textcolor[rgb]{0.8,0,0}{\ding{55}}}

\definecolor{categorygreen}{RGB}{232, 245, 233}
\definecolor{categoryorange}{RGB}{255, 243, 224}
\definecolor{categorygray}{RGB}{245, 245, 245}
\definecolor{categorypurple}{RGB}{243, 229, 245}
\definecolor{oursblue}{RGB}{232, 240, 254}

\newtcbox{\hlprimarytab}{on line, rounded corners, box align=base,
  colback=green!10, colframe=white, size=fbox, arc=3pt,
  before upper=\strut, top=-2pt, bottom=-4pt, left=-2pt, right=-2pt, boxrule=0pt}

\newtcbox{\hlsecondarytab}{on line, rounded corners, box align=base,
  colback=red!10, colframe=white, size=fbox, arc=3pt,
  before upper=\strut, top=-2pt, bottom=-4pt, left=-2pt, right=-2pt, boxrule=0pt}

\newcommand{\dashifted}{\raisebox{0.5\depth}{\tiny$\downarrow$}}
\newcommand{\uashifted}{\raisebox{0.5\depth}{\tiny$\uparrow$}}
\newcommand{\dar}[1]{{\raisebox{0.6ex}{\tiny\hlsecondarytab{\dashifted\,#1\%}}}}
\newcommand{\uar}[1]{{\raisebox{0.6ex}{\tiny\hlprimarytab{\uashifted\,#1\%}}}}

\title{TouchAnything: A Dataset and Framework for Bimanual Tactile Estimation from Egocentric Video}
\renewcommand{\authorfont}{\fontsize{8.5}{10}\selectfont}

\author[1]{Jianyi Zhou}
\author[1]{Ziteng Gao}
\author[1]{Feiyang Hong}
\author[1]{Zirui Liu}
\author[1]{Guannan Zhang}
\author[1]{Weisheng Dai}
\author[2]{Ruichen Zhen}
\author[3]{\\Chuqiao Lyu}
\author[2]{Haotian Wu}
\author[2]{Yinian Mao}
\author[1]{Xushi Wang}
\author[1]{Yuxiang Jiang}
\author[3]{Wenbo Ding}
\author[1\text{\ding{41}}]{Shuo Yang}

\renewcommand\affiliation[2][]{\addtolist[#1]{#2}{\affiliationlist}{\affiliationformat}{\\}}
\affiliation[1]{Harbin Institute of Technology, Shenzhen}
\affiliation[2]{Meituan Academy of Robotics}
\affiliation[3]{Tsinghua Shenzhen International Graduate School, Tsinghua University}

\abstract{
Egocentric human video data, which captures rich human-environment interactions and can be collected at scale, has become a key driver of embodied intelligence research. However, existing egocentric datasets typically lack tactile sensing, a critical modality that provides direct cues about contact, force, and pressure in human-object interaction. Without such signals, models struggle to learn physically grounded representations of real-world interaction dynamics. 
While tactile sensors provide these cues, deploying high-quality tactile hardware at scale remains expensive and cumbersome. This raises a central question: can tactile feedback be inferred directly from visual observations, enabling scalable tactile supervision for egocentric video data and supporting physically grounded embodied learning?
To enable research in this direction, we introduce \textbf{EgoTouch}, a large-scale multi-view egocentric dataset with dense tactile supervision for bimanual hand-object interaction. EgoTouch comprises 208 manipulation tasks spanning 1,891 episodes in diverse indoor and outdoor environments, with synchronized multi-view RGB (head-mounted egocentric and dual wrist-mounted cameras), bimanual 3D hand pose, and continuous pressure maps from wearable tactile sensors. Building on EgoTouch, we introduce \textbf{TouchAnything}, a baseline multi-view vision-to-touch prediction framework that uses the egocentric view as the primary input and flexibly leverages available wrist-mounted views at inference time.
Experiments show that incorporating wrist-mounted views generally improves tactile prediction over egocentric-only input, achieving up to 5.0\% relative improvement in Contact IoU and 6.1\% relative improvement in Volumetric IoU. We will publicly release the dataset, code, and benchmark.
}
\date{\today}
\correspondence{\email{shuoyang@hit.edu.cn}}

\checkdata[Project Page]{\url{https://jianyi2004.github.io/TouchAnything-Website/}}

\begin{document}
\begin{CJK*}{UTF8}{gbsn}

\maketitle

\begin{figure}[t]
\centering
\includegraphics[width=\textwidth]{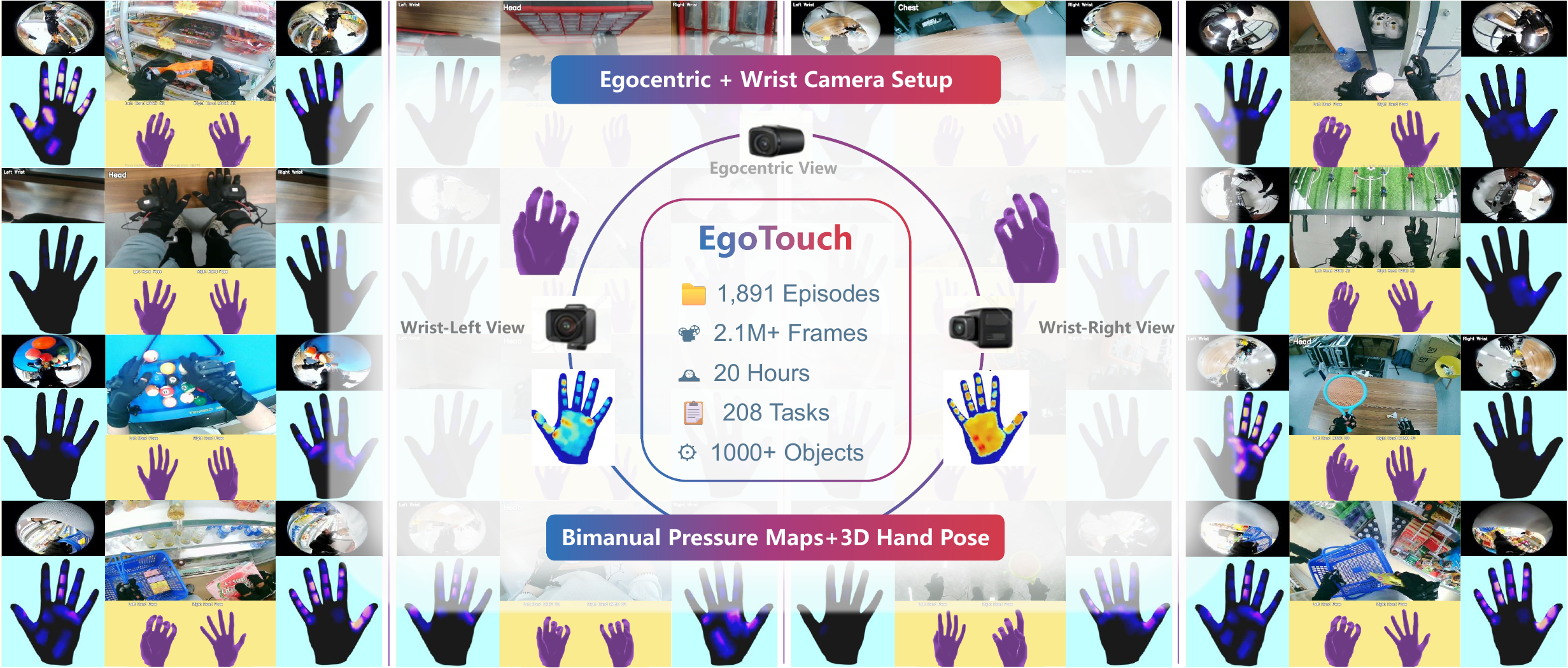}
\caption{EgoTouch combines egocentric and wrist-mounted views with synchronized 3D hand pose and dense tactile pressure maps, providing complementary visual evidence for learning contact-aware interactions.}
\label{fig:teaser}
\end{figure}


\section{Introduction}
\label{sec:intro}


Egocentric video datasets have become a key source of scalable supervision for embodied intelligence, as they are relatively easy to collect, capture natural human-environment interactions, and cover diverse real-world manipulation behaviors~\citep{grauman2022ego4d,zheng2026egoscalescalingdexterousmanipulation,hoque2026egodexlearningdexterousmanipulation,DBLP:journals/corr/abs-2603-22264,DBLP:journals/corr/abs-2601-12993,DBLP:journals/ijcv/GraumanWTKMAABBBBCCCCCD25,DBLP:journals/corr/abs-2602-06949,DBLP:journals/corr/abs-2507-12440}.
However, despite this progress, a critical modality remains largely missing: tactile sensing. While visual observations capture appearance and motion cues, they do not directly reveal the physical signals underlying successful manipulation, such as contact force and pressure distribution. Without access to such signals, embodied models lack direct supervision of the physical interaction dynamics that govern real-world manipulation~\citep{DBLP:journals/ral/YangHLTLSP23,DBLP:conf/iros/LinCCZL23}, limiting their ability to develop a deeper understanding of the physical world~\citep{DBLP:journals/ar/YamaguchiA19,DBLP:journals/ras/SuomalainenKK22}.
Although tactile sensors can provide this information, collecting large-scale tactile data with high-quality hardware is expensive, intrusive, and difficult to scale. This creates a fundamental bottleneck: the availability of large-scale visual data contrasts sharply with the scarcity of tactile supervision.
This gap raises a key question: can tactile signals be inferred directly from visual observations, enabling scalable tactile supervision for large-scale egocentric data and supporting interaction-aware embodied learning?

Vision-to-touch prediction has emerged as a promising direction~\citep{yang2022pressurevision,li2019visgel,DBLP:conf/nips/YangMZZYO22}, but progress remains fundamentally constrained by data. Existing datasets either rely on single-view capture~\citep{yang2022pressurevision} or focus on relatively narrow interaction settings such as hand-surface contact or single-finger pressing~\citep{grady2024egopressure}. As a result, they provide limited support for studying tactile prediction in realistic bimanual hand-object interactions, which involve diverse manipulation contexts and frequent occlusion of hand-object contact regions.
This occlusion is a central challenge in egocentric vision-to-touch prediction: contact regions are often hidden by the hand itself or the manipulated object, making tactile signals only partially observable from the head-mounted view. This missing contact evidence introduces substantial ambiguity, especially in complex manipulation scenarios.

To address these challenges, we introduce \textbf{EgoTouch} (as shown in Figure~\ref{fig:teaser}), a large-scale multi-view egocentric dataset with dense tactile supervision for bimanual hand-object interaction. EgoTouch pairs egocentric and wrist-mounted videos with synchronized tactile pressure maps and bimanual hand pose, providing complementary views of realistic hand-object interactions. It contains 208 manipulation tasks, 1,891 episodes, and over 20 hours of interaction data across diverse indoor and outdoor environments, supporting cross-modal learning of physical interaction dynamics.

Building on EgoTouch, we establish \textbf{TouchAnything}, a baseline vision-to-touch prediction model that supports flexible inference with single or multiple camera views. A key design choice is view dropout: during training, wrist camera views are randomly dropped, forcing the model to learn robust representations that work with any subset of available views at inference time. This enables deployment in settings where only an egocentric camera is available, while gracefully leveraging additional views when present. Experiments show that adding wrist-mounted views improves tactile prediction on both seen and unseen objects: the full multi-view setting improves Contact IoU from 0.4792 to 0.5030 and Volumetric IoU from 0.4311 to 0.4575 on seen objects, and improves Contact IoU from 0.4396 to 0.4496 and Volumetric IoU from 0.3743 to 0.3852 on unseen objects.

In summary, our contributions are:
\begin{enumerate}[leftmargin=*,itemsep=2pt]
\item We introduce \textbf{EgoTouch}, a large-scale multi-view egocentric dataset for bimanual hand-object interaction, comprising 208 tasks, 1,891 episodes, synchronized RGB videos from one head-mounted camera and two wrist-mounted cameras, bimanual 3D hand pose, and dense continuous pressure maps across diverse environments.

\item We establish \textbf{a multi-view vision-to-touch benchmark} on EgoTouch, with evaluation protocols for seen and unseen objects and different camera-view configurations, enabling systematic analysis of how complementary wrist views affect tactile prediction.

\item We propose \textbf{TouchAnything}, a baseline vision-to-touch model with cross-view fusion and view dropout training, supporting flexible inference with egocentric-only or multi-view inputs and improving tactile prediction when wrist views are available.
\end{enumerate}


\begin{table}[t]
\caption{Comparison of EgoTouch with existing hand interaction and tactile datasets. EgoTouch is the first to jointly provide multi-view video, bimanual hand pose, and real dense pressure data across diverse scenes.}
\label{tab:dataset_comparison}
\centering
\small
\setlength{\tabcolsep}{4pt}
\begin{tabular}{lccccccc}
\toprule
Dataset & In-the-wild & Hand Pose & Contact & Wrist Views & Hands & Objects  & Frames \\
\midrule
GRAB~\citep{grab} & \xmark & MoCap & Analytical & \xmark & Biman. & 51 & 1.6M \\
ContactDB~\citep{contactDB} & \xmark & \xmark & Thermal & \xmark & Biman. & 50 & 375k \\
ARCTIC~\citep{arctic} & \xmark & MoCap & Analytical & \xmark & Biman. & 11 & 2.1M \\
OakInk~\citep{yang2022oakink} & \xmark & MoCap & Analytical & \xmark & Single & 100 & 230k \\
DexYCB~\citep{dexycb} & \xmark & Est. & Analytical & \xmark & Single & 20 & 582k \\
ActionSense~\citep{delpreto2022actionsense} & \xmark & Glove & Pressure & \xmark & Biman. & 21 & 521k \\
HOI4D~\citep{liu2022hoi4d} & \xmark & Est. & \xmark & \xmark & Single & 800 & 2.4M \\
EgoPressure~\citep{grady2024egopressure} & \xmark & Est. & Pressure & \xmark & Single & 31 & 4.3M \\
EgoDex~\citep{hoque2026egodexlearningdexterousmanipulation} & \xmark & Est. & Analytical & \xmark & Biman. & 500 & 90M \\
OpenTouch~\citep{song2025opentouchbringingfullhandtouch} & \cmark & Glove & Pressure & \xmark & Single & ~800 & $\sim$ 500k \\
\midrule
\textbf{EgoTouch (Ours)} & \cmark & \textbf{Glove + Est.} & \textbf{Pressure} & \cmark & \textbf{Biman.} & \textbf{$\sim$1000} & \textbf{2.1M} \\
\bottomrule
\end{tabular}
\end{table}

\section{EgoTouch Dataset}
\label{sec:dataset}
The EgoTouch dataset contains 20 hours of multi-view egocentric video collected at 30 Hz, comprising 1,891 episodes across 208 diverse manipulation tasks. This amounts to approximately 2.1 million frames covering over 1,000 objects in both indoor and outdoor environments. The dataset provides rich and structured annotations, including synchronized multi-view RGB videos, bimanual 3D hand pose, and dense tactile pressure maps for both hands. All modalities are temporally aligned at the frame level to enable precise cross-modal learning. We compare EgoTouch with existing hand-object interaction and tactile datasets in Table~\ref{tab:dataset_comparison}.

\subsection{Data Collection Setup}
\label{sec:data_collection}
\begin{figure}[t]
\centering
\includegraphics[width=\textwidth]{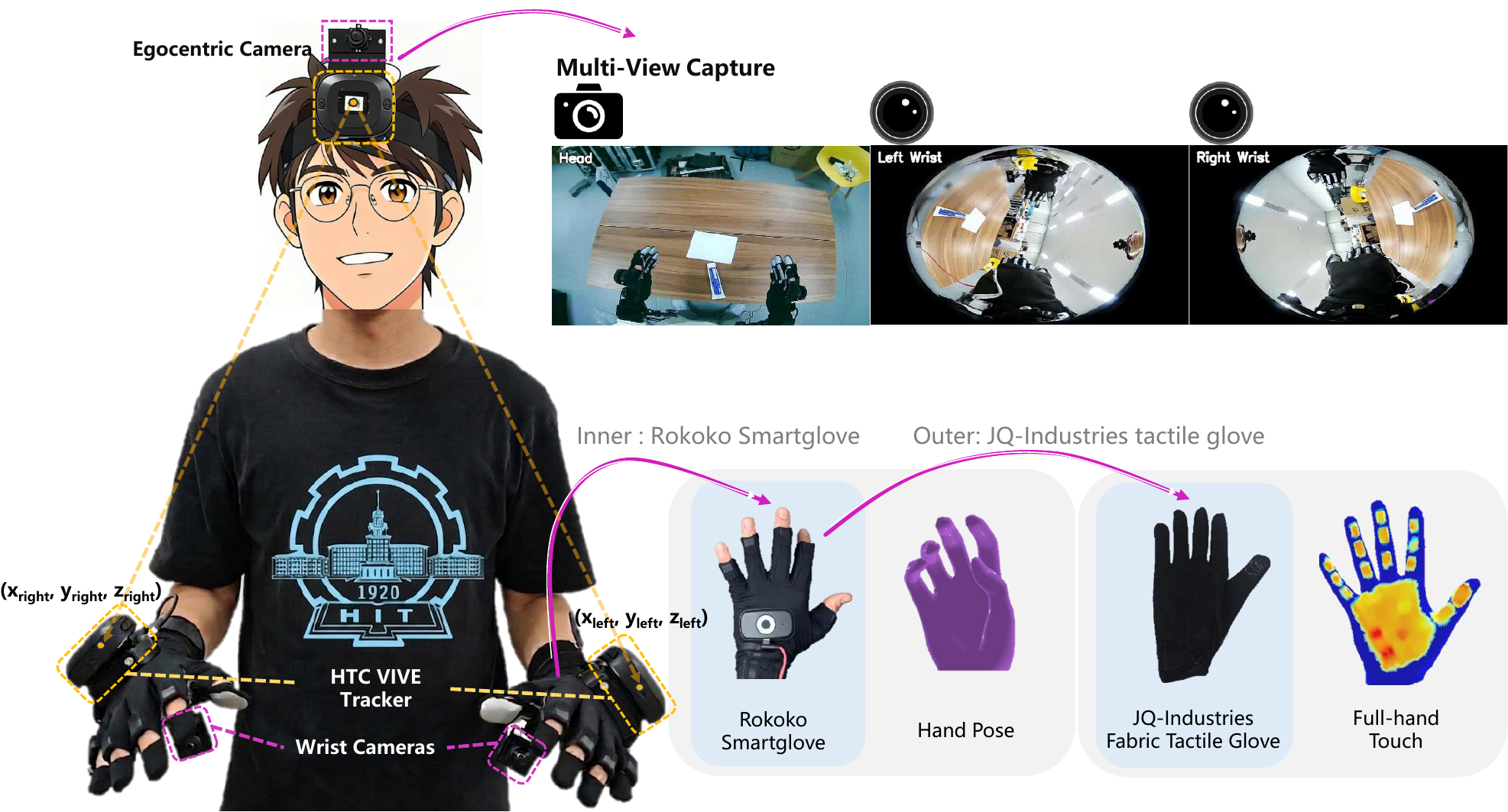}
\caption{Data collection setup and example multi-modal data. The participant wears a head-mounted egocentric camera, two wrist-mounted cameras, and pressure-sensing gloves. All modalities are temporally synchronized.}
\label{fig:collection_setup}
\end{figure}

\begin{figure}[t]
\centering
\includegraphics[width=\textwidth]{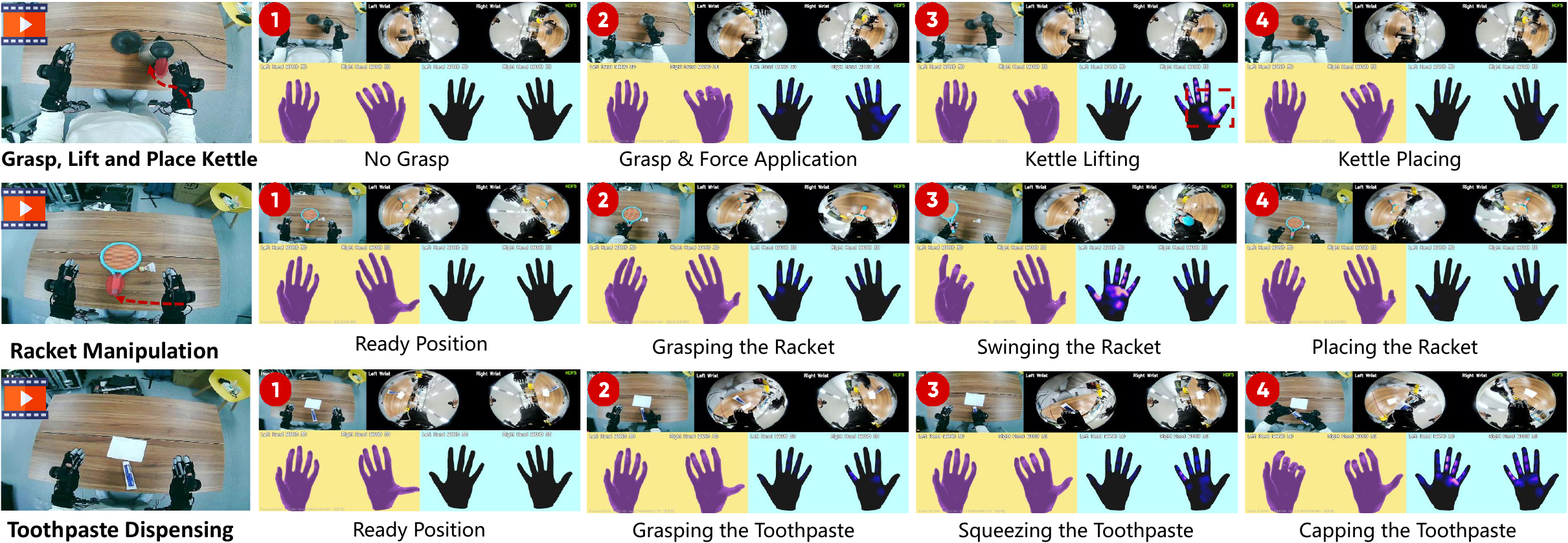}
\caption{Example data from EgoTouch demonstrates that hardware-based tactile sensing and pose tracking reveal critical
force, contact, and motion cues that vision alone cannot capture.}
\label{fig:example_data}
\end{figure}

EgoTouch is collected with a synchronized wearable capture system that records complementary visual, kinematic, and tactile signals during natural bimanual manipulation (Figure~\ref{fig:collection_setup}). The setup includes a head-mounted RGB camera for global egocentric context, two wrist-mounted RGB cameras for close-up observations of hand-object contact regions, Rokoko motion-capture gloves for bimanual 3D hand pose, custom pressure-sensing gloves with $16 \times 16$ tactile arrays on each palm, and HTC Vive Trackers for 6-DoF head and wrist localization. All streams are synchronized onto a shared 30Hz timeline using timestamps and latest valid sensor snapshots. This produces frame-level synchronized multi-view RGB, hand pose, tactile pressure maps, and tracker poses. Figure~\ref{fig:example_data} shows representative synchronized observations from the dataset. Additional hardware specifications, acquisition details, and synchronization strategy are provided in Appendix~\ref{sec:appendix_data_collection}.

\subsection{Data Modalities}
\label{sec:modalities}

Each frame in EgoTouch is organized on a synchronized 30Hz timeline and contains the following modalities:

\begin{itemize}[leftmargin=*,itemsep=2pt]

\item \textbf{Multi-view RGB videos.}
The dataset provides three egocentric RGB views: a head-mounted camera capturing the global manipulation scene ($V^h \in \mathbb{R}^{640 \times 480 \times 3}$), and two wrist-mounted fisheye cameras ($V^{wL}, V^{wR} \in \mathbb{R}^{640 \times 480 \times 3}$) providing close-up observations of hand-object contact regions during bimanual interactions.

\item \textbf{Bimanual 3D hand pose.}
Hand kinematics are represented by 42 three-dimensional joints ($\mathbf{P} \in \mathbb{R}^{42 \times 3}$), including wrist, finger, and fingertip joints for both hands.

\item \textbf{Tactile pressure maps.}
Dense tactile feedback is recorded as bilateral $16 \times 16$ raw pressure arrays
($\mathbf{M}_{raw} \in \mathbb{R}^{2 \times 16 \times 16}$), which are normalized and remapped into canonical
$21 \times 21$ hand-shaped grids ($\mathbf{M} \in \mathbb{R}^{2 \times 21 \times 21}$) for training.
Details are provided in Appendix~\ref{sec:appendix_tactile_preprocessing}.

\item \textbf{Tracker poses and metadata.}
Each frame additionally includes 6-DoF poses from HTC Vive Trackers mounted on the head and wrists, together with metadata annotations such as task category, object category, scene description, and environment type.

\end{itemize}

All modalities are temporally aligned at the frame level to support cross-modal learning of physical interaction dynamics.

\subsection{Task Taxonomy}
\label{sec:tasks}

\begin{figure}[t]
\centering
\includegraphics[width=\textwidth]{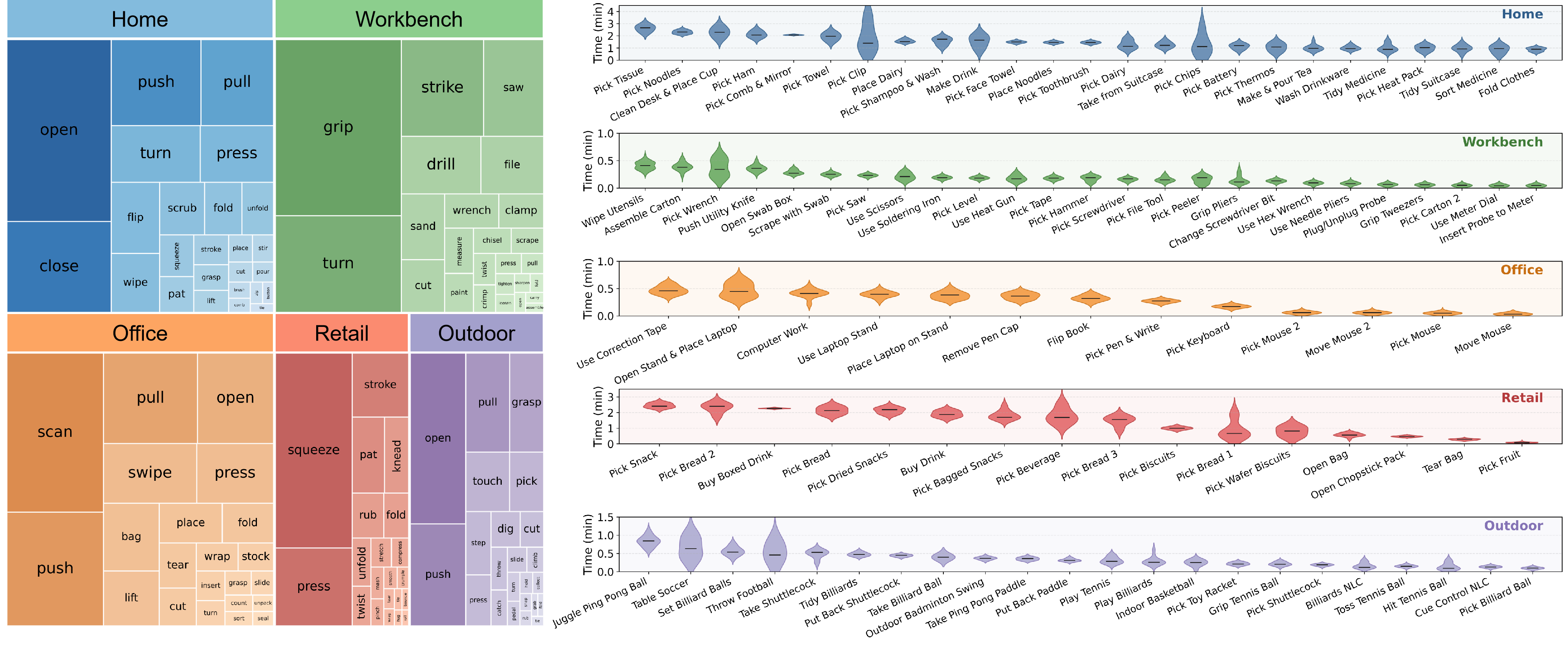}
\caption{Dataset statistics and analysis. EgoTouch exhibits broad task coverage across five environment categories.}
\label{fig:statistics}
\end{figure}

The 208 tasks in EgoTouch are grouped into five environment-based categories that capture diverse real-world interaction patterns (Figure~\ref{fig:statistics}):

\begin{itemize}[leftmargin=*,itemsep=2pt]

\item \textbf{Home}: Everyday household interactions such as opening containers, pushing and pulling objects, pressing switches, wiping surfaces, folding clothes, and handling daily items.

\item \textbf{Workbench}: Tool-based manipulation tasks including gripping and turning tools, sawing, drilling, sanding, cutting, clamping, and precision assembly interactions.

\item \textbf{Office}: Workspace activities such as  swiping cards, typing on keyboards, operating office tools, and manipulating books or stationery.

\item \textbf{Retail}: Consumer interaction behaviors including squeezing products, pressing packaged items, folding goods, opening bags, and handling snacks or beverages.

\item \textbf{Outdoor}: Dynamic open-environment interactions including ball games, racket sports, outdoor object handling, and other full-body coordinated manipulation activities.

\end{itemize}

\section{METHOD: Tactile Prediction Framework}
\label{sec:method}

We introduce a baseline framework for multi-view tactile prediction that maps visual observations and hand pose to dense bimanual pressure maps. The framework is designed to (1) leverage complementary viewpoints to address occlusion, and (2) support flexible inference under missing views.

\subsection{Problem Formulation}
\label{sec:formulation}

Given a video clip of $T$ frames from a subset of views $\mathcal{V} \subseteq \{V^{ego}, V^{wL}, V^{wR}\}$ and the corresponding bimanual hand pose sequence $\mathbf{P} \in \mathbb{R}^{T \times 42 \times 3}$, our goal is to predict bilateral tactile maps $\hat{\mathbf{M}} \in \mathbb{R}^{T \times 2 \times 21 \times 21}$ at each timestep, where the tactile maps are represented in a canonical $21 \times 21$ hand-shaped grid after preprocessing and spatial remapping of the raw tactile sensor layout (Appendix~\ref{sec:appendix_tactile_preprocessing}).

\subsection{Multi-View Tactile Prediction Framework}

Our framework consists of a shared visual encoder, a cross-view fusion module, a pose-aware fusion mechanism, and a tactile decoder, enabling joint modeling of appearance, geometry, and motion cues.

Each input view is processed independently using a shared visual backbone, followed by a learnable view embedding to encode camera identity. This allows the model to distinguish between egocentric and wrist-mounted perspectives while maintaining parameter efficiency. To integrate information across views, we apply a lightweight cross-view attention module over view-level features, enabling complementary reasoning across viewpoints. For example, wrist views can provide contact information that is occluded in the egocentric view. The fused representation is further aggregated using a gated mechanism that dynamically weighs the contribution of each view, ensuring robustness to missing or unreliable inputs.

\begin{figure}[t]
\centering
\includegraphics[width=\textwidth]{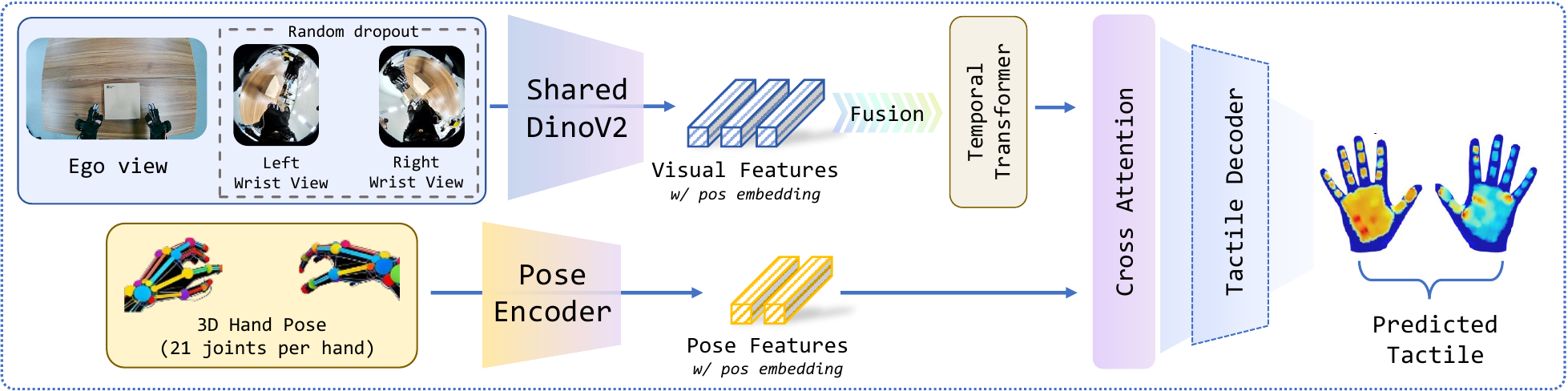}
\caption{Architecture of the multi-view tactile prediction model. A shared backbone encodes each view with view embeddings. Cross-view attention and gated fusion produce unified visual features, which are combined with hand pose through pose-aware fusion and decoded into bilateral pressure maps.}
\label{fig:architecture}
\end{figure}

To incorporate geometric information, we encode bimanual hand pose into joint-level features and fuse them with visual representations via cross-attention. Each joint attends to the most relevant visual regions, enabling spatially grounded reasoning about contact. This design allows the model to associate tactile signals with specific hand regions, which is critical for predicting structured pressure distributions. Temporal dependencies are modeled using a lightweight temporal module applied to the fused features, capturing interaction dynamics such as grasping and sliding.

The fused joint-level features are decoded into bilateral pressure maps for both hands. Each hand is predicted independently as a $21 \times 21$ pressure grid, representing normalized contact intensity. This formulation enables dense spatial supervision beyond binary contact prediction.

\subsection{View Dropout and Training Objective}

To support varying sensor configurations, we employ a view dropout strategy during training. The egocentric view is always retained, while wrist views are randomly dropped. This exposes the model to different input combinations and enables flexible inference at test time without architectural changes. The model can operate with only egocentric input, while benefiting from additional views when available.

The model is trained using a weighted regression loss that combines pixel-wise reconstruction with spatial regularization:
\begin{equation}
\mathcal{L} = \lambda_{mse}\mathcal{L}_{MSE} + \lambda_{l1}\mathcal{L}_{L1} + \lambda_{tv}\mathcal{L}_{TV}(\hat{\mathbf{M}})
\end{equation}
where $\mathcal{L}_{MSE}$ and $\mathcal{L}_{L1}$ measure pressure reconstruction error, and $\mathcal{L}_{TV}$ encourages spatial smoothness. To address the sparsity of tactile maps and prevent the model from collapsing to all-zero predictions, we apply higher loss weights to contact regions where pressure exceeds a threshold (0.1). We use $\lambda_{mse}=1.0$, $\lambda_{l1}=0.5$, $\lambda_{tv}=0.01$, and a contact-region weight of 3.0.

\section{Experiments}
\label{sec:experiments}

\subsection{Experimental Setup}
\label{sec:impl}
\textbf{Dataset split.}
We split the dataset into training, validation, and test sets with a ratio of 80\% / 10\% / 10\% at the episode level to avoid temporal data leakage. The test set is further divided into \textit{seen-object} and \textit{unseen-object} subsets to evaluate generalization to novel object instances.
All splits cover a diverse set of interaction scenarios, ensuring variation in objects, tasks, and environments. This setup enables evaluation under both in-distribution and out-of-distribution settings.

\textbf{Evaluation metric.}
 We define the following benchmark tasks on EgoTouch:
1) Tactile prediction. Given multi-view video and hand pose, predict the bilateral pressure map. This is the primary benchmark.
2) Contact detection. A binarized version: predict whether each region is in contact (pressure $> \tau$). Derived from the tactile prediction output.

We evaluate with the following metrics, following PressureVision~\citep{yang2022pressurevision}:

\begin{itemize}[leftmargin=*,itemsep=1pt]
\item \textbf{Temporal Accuracy} $\uparrow$: Evaluates the temporal accuracy of contact onset and termination. If any contact is present in the estimated and ground truth contact maps, the frame is marked as in contact. A frame is marked correct if the presence of contact is consistent in estimated and ground truth frames.

\item \textbf{Contact IoU} $\uparrow$: Evaluates the spatial and temporal accuracy of estimated contact by computing the intersection over union (IoU) between the binary contact images. This metric does not consider the magnitude of the estimated pressure, and is an upper bound on Volumetric IoU.

\item \textbf{Volumetric IoU} $\uparrow$: Extends Contact IoU to evaluate the magnitudes of pressure estimates in addition to their spatial and temporal accuracy. Each 2D pressure image is converted into a 3D ``pressure volume'', where the height of the volume is equal to the amount of pressure at that pixel. The Volumetric IoU is calculated as:
\begin{equation}
IoU_{vol} = \frac{\sum^{i,j} min(P_{i,j}, \hat{P}_{i,j})}{\sum^{i,j} max(P_{i,j}, \hat{P}_{i,j})}
\end{equation}
where $P_{i,j}$ is the ground truth pressure at pixel $(i,j)$ and $\hat{P}_{i,j}$ is the predicted pressure.

\item \textbf{MAE} $\downarrow$: Mean absolute error over normalized pressure values. We calculate MAE over each pixel. As most of the dataset pressure images consist of zeros, these numbers are close to zero.
\end{itemize}
\textbf{Implementation.}
We implement TouchAnything in PyTorch and train it on NVIDIA GPUs with distributed data parallelism. The visual encoder is a frozen DINOv2-Base (ViT-B/14) backbone initialized from a pretrained checkpoint. Each training sample consists of a clip of $T=8$ frames sampled with a frame interval of 2 from three synchronized RGB views, including one egocentric view and two wrist-mounted views. All frames are resized to $224 \times 224$ and paired with 42 3D hand joints and bilateral $21 \times 21$ tactile maps normalized to $[0,1]$. The model is trained for 25 epochs using AdamW with a learning rate of $5 \times 10^{-5}$, weight decay of 0.05, and a cosine learning-rate schedule with 10 warmup epochs. We optimize a contact-aware weighted reconstruction objective that combines MSE and L1 losses with a total-variation regularizer; tactile cells with pressure larger than 0.1 are assigned a weight of 3.0 to mitigate the sparsity of contact signals and discourage trivial all-zero predictions. During training, each wrist view is independently dropped with probability $p=0.3$, while the egocentric view is always retained. This view-dropout strategy enables flexible inference with any available subset of camera views.

\subsection{Main Results}
\label{sec:main_results}

\begin{table*}[t]
\caption{Multi-view tactile prediction across five diverse scenarios. All methods use the same architecture and training recipe. Arrows ($\uparrow/\downarrow$) show relative percentage change vs. the Ego-only baseline.}
\label{tab:main_results}
\centering
\setlength{\tabcolsep}{3pt}
\resizebox{\textwidth}{!}{%
\begin{tabular}{c|llll|llll}
\toprule
 & \multicolumn{4}{c|}{\textit{Seen Objects}} & \multicolumn{4}{c}{\textit{Unseen Objects}} \\
\cmidrule(lr){2-5} \cmidrule(lr){6-9}
\textbf{Method} & \textbf{T.Acc} & \textbf{C.IoU} & \textbf{V.IoU} & \textbf{MAE} & \textbf{T.Acc} & \textbf{C.IoU} & \textbf{V.IoU} & \textbf{MAE} \\
\midrule
\rowcolor{categorygreen} \multicolumn{9}{l}{\normalsize\textit{Home Scenario}} \\
Ego-only & 0.8532 & 0.4869 & 0.4361 & 0.0470 & 0.8175 & 0.4250 & 0.3671 & 0.0604 \\
Ego + wL & 0.8659\uar{1.5} & 0.5074\uar{4.2} & 0.4598\uar{5.4} & 0.0453\dar{3.6} & 0.8278\uar{1.3} & 0.4375\uar{2.9} & 0.3779\uar{2.9} & 0.0591\dar{2.2} \\
Ego + wR & 0.8658\uar{1.5} & 0.5072\uar{4.2} & 0.4597\uar{5.4} & 0.0453\dar{3.6} & 0.8276\uar{1.2} & 0.4372\uar{2.9} & 0.3776\uar{2.9} & 0.0592\dar{2.0} \\
\textbf{Ego + wL + wR} & \textbf{0.8659\uar{1.5}} & \textbf{0.5075\uar{4.2}} & \textbf{0.4598\uar{5.4}} & \textbf{0.0452\dar{3.8}} & \textbf{0.8267\uar{1.1}} & \textbf{0.4371\uar{2.8}} & \textbf{0.3775\uar{2.8}} & \textbf{0.0590\dar{2.3}} \\
\midrule
\rowcolor{categoryorange} \multicolumn{9}{l}{\normalsize\textit{Workbench Scenario}} \\
Ego-only & 0.7954 & 0.4131 & 0.3747 & 0.0420 & 0.8096 & 0.2932 & 0.2801 & 0.0599 \\
Ego + wL & 0.8250\uar{3.7} & 0.4422\uar{7.0} & 0.4080\uar{8.9} & 0.0395\dar{6.0} & 0.7525\dar{7.1} & 0.3016\uar{2.9} & 0.2841\uar{1.4} & 0.0579\dar{3.3} \\
Ego + wR & 0.8242\uar{3.6} & 0.4408\uar{6.7} & 0.4070\uar{8.6} & 0.0395\dar{6.0} & 0.7539\dar{6.9} & 0.3017\uar{2.9} & 0.2840\uar{1.4} & 0.0579\dar{3.3} \\
\textbf{Ego + wL + wR} & \textbf{0.8241\uar{3.6}} & \textbf{0.4414\uar{6.9}} & \textbf{0.4078\uar{8.8}} & \textbf{0.0395\dar{6.0}} & \textbf{0.7538\dar{6.9}} & \textbf{0.3015\uar{2.8}} & \textbf{0.2840\uar{1.4}} & \textbf{0.0578\dar{3.5}} \\
\midrule
\rowcolor{categorygray} \multicolumn{9}{l}{\normalsize\textit{Office Scenario}} \\
Ego-only & 0.8770 & 0.4918 & 0.4501 & 0.0436 & 0.8901 & 0.5743 & 0.4511 & 0.0632 \\
Ego + wL & 0.8990\uar{2.5} & 0.5249\uar{6.7} & 0.4889\uar{8.6} & 0.0401\dar{8.0} & 0.9035\uar{1.5} & 0.5718\dar{0.4} & 0.4673\uar{3.6} & 0.0615\dar{2.7} \\
Ego + wR & 0.9001\uar{2.6} & 0.5250\uar{6.8} & 0.4891\uar{8.7} & 0.0400\dar{8.3} & 0.9035\uar{1.5} & 0.5718\dar{0.4} & 0.4672\uar{3.6} & 0.0615\dar{2.7} \\
\textbf{Ego + wL + wR} & \textbf{0.9003\uar{2.7}} & \textbf{0.5256\uar{6.9}} & \textbf{0.4895\uar{8.8}} & \textbf{0.0399\dar{8.5}} & \textbf{0.9035\uar{1.5}} & \textbf{0.5718\dar{0.4}} & \textbf{0.4672\uar{3.6}} & \textbf{0.0615\dar{2.7}} \\
\midrule
\rowcolor{categorypurple} \multicolumn{9}{l}{\normalsize\textit{Retail Scenario}} \\
Ego-only & 0.7907 & 0.4454 & 0.4112 & 0.0411 & 0.5841 & 0.3935 & 0.3636 & 0.0313 \\
Ego + wL & 0.8334\uar{5.4} & 0.4901\uar{10.0} & 0.4414\uar{7.3} & 0.0392\dar{4.6} & 0.6591\uar{12.8} & 0.4011\uar{1.9} & 0.3811\uar{4.8} & 0.0310\dar{1.0} \\
Ego + wR & 0.8241\uar{4.2} & 0.4851\uar{8.9} & 0.4390\uar{6.8} & 0.0395\dar{3.9} & 0.6670\uar{14.2} & 0.4008\uar{1.9} & 0.3812\uar{4.8} & 0.0310\dar{1.0} \\
\textbf{Ego + wL + wR} & \textbf{0.8316\uar{5.2}} & \textbf{0.4900\uar{10.0}} & \textbf{0.4411\uar{7.3}} & \textbf{0.0391\dar{4.9}} & \textbf{0.6591\uar{12.8}} & \textbf{0.4007\uar{1.8}} & \textbf{0.3810\uar{4.8}} & \textbf{0.0309\dar{1.3}} \\
\midrule
\rowcolor{oursblue} \multicolumn{9}{l}{\normalsize\textit{Outdoor Scenario}} \\
Ego-only & 0.7905 & 0.5339 & 0.4769 & 0.0456 & 0.9745 & 0.4879 & 0.3750 & 0.1071 \\
Ego + wL & 0.8060\uar{2.0} & 0.5533\uar{3.6} & 0.5002\uar{4.9} & 0.0439\dar{3.7} & 0.9628\dar{1.2} & 0.4978\uar{2.0} & 0.3850\uar{2.7} & 0.1043\dar{2.6} \\
Ego + wR & 0.8062\uar{2.0} & 0.5530\uar{3.6} & 0.5001\uar{4.9} & 0.0439\dar{3.7} & 0.9628\dar{1.2} & 0.4986\uar{2.2} & 0.3854\uar{2.8} & 0.1043\dar{2.6} \\
\textbf{Ego + wL + wR} & \textbf{0.8061\uar{2.0}} & \textbf{0.5534\uar{3.7}} & \textbf{0.5002\uar{4.9}} & \textbf{0.0439\dar{3.7}} & \textbf{0.9628\dar{1.2}} & \textbf{0.4981\uar{2.1}} & \textbf{0.3851\uar{2.7}} & \textbf{0.1043\dar{2.6}} \\
\midrule
\rowcolor{yellow!25} \multicolumn{9}{l}{\normalsize\textit{Overall (All Scenarios)}} \\
Ego-only & 0.8393 & 0.4792 & 0.4311 & 0.0456 & 0.8271 & 0.4396 & 0.3743 & 0.0615 \\
Ego + wL & 0.8567\uar{2.1} & 0.5030\uar{5.0} & 0.4575\uar{6.1} & 0.0437\dar{4.2} & 0.8355\uar{1.0} & 0.4499\uar{2.3} & 0.3856\uar{3.0} & 0.0601\dar{2.3} \\
Ego + wR & 0.8561\uar{2.0} & 0.5024\uar{4.8} & 0.4572\uar{6.1} & 0.0437\dar{4.2} & 0.8355\uar{1.0} & 0.4497\uar{2.3} & 0.3854\uar{3.0} & 0.0602\dar{2.1} \\
\textbf{Ego + wL + wR} & \textbf{0.8566\uar{2.1}} & \textbf{0.5030\uar{5.0}} & \textbf{0.4575\uar{6.1}} & \textbf{0.0436\dar{4.4}} & \textbf{0.8347\uar{0.9}} & \textbf{0.4496\uar{2.3}} & \textbf{0.3852\uar{2.9}} & \textbf{0.0601\dar{2.3}} \\
\bottomrule
\end{tabular}%
}
\end{table*}
Table~\ref{tab:main_results} shows that wrist-mounted views generally improve tactile prediction over the egocentric-only baseline, particularly for Contact IoU, Volumetric IoU, and MAE. Overall, Ego + wL + wR improves Contact IoU from 0.4792 to 0.5030 and Volumetric IoU from 0.4311 to 0.4575 on seen objects, and from 0.4396 to 0.4496 and 0.3743 to 0.3852 on unseen objects, respectively. These gains indicate that wrist views provide complementary contact-region evidence, especially for localizing contact and estimating pressure magnitude under egocentric occlusion.

Temporal Accuracy changes more modestly and varies across scenarios, suggesting that wrist views mainly help resolve where and how strongly contact occurs rather than simply whether contact occurs. We also find that a single wrist view already captures much of the complementary evidence, likely because the fisheye cameras cover a broad interaction region and can sometimes observe details of the opposite hand. Thus, the main benefit comes from adding at least one contact-aware viewpoint, while the second wrist view provides additional gains mainly under stronger bimanual occlusion.

\subsection{Ablation Studies}
\label{sec:ablations}
\textbf{View dropout is important.} As shown in Table~\ref{tab:ablation_dropout}, training without view dropout leads to a substantial performance drop when wrist views are unavailable at inference time, indicating that the model relies heavily on the full multi-view setting. In contrast, incorporating view dropout significantly improves robustness: the relative performance drop ($\Delta$ V) between all-view and ego-only inference improves from -27.20\% to -5.78\% on the seen-object split.

Importantly, view dropout enables strong generalization to partial-view inputs while maintaining competitive full multi-view performance. This suggests that exposure to varying view combinations during training encourages the model to learn complementary representations across views, rather than relying on a fixed camera configuration. As a result, the model can flexibly operate under different deployment conditions where some views may be missing or occluded.

\begin{table}[t]
\caption{View-dropout ablation on the seen-object split. View dropout ($p{=}0.3$) improves robustness to missing wrist views; Ego + wL/wR reports the average performance over the two single-wrist configurations; $\Delta$V denotes the relative V.IoU drop from all-view inference under the same training strategy.}
\label{tab:ablation_dropout}
\centering
\small
\setlength{\tabcolsep}{2.8pt}
\resizebox{\linewidth}{!}{%
\begin{tabular}{l|cccc|cccc|cccc}
\toprule
\multirow{2}{*}{\textbf{Training}} & \multicolumn{4}{c|}{\textbf{Ego only}} & \multicolumn{4}{c|}{\textbf{Ego + wL/wR}} & \multicolumn{4}{c}{\textbf{All views}} \\
\cmidrule(lr){2-5} \cmidrule(lr){6-9} \cmidrule(lr){10-13}
& \textbf{C.IoU} & \textbf{V.IoU} & \textbf{MAE} & \textbf{$\Delta$V} & \textbf{C.IoU} & \textbf{V.IoU} & \textbf{MAE} & \textbf{$\Delta$V} & \textbf{C.IoU} & \textbf{V.IoU} & \textbf{MAE} & \textbf{$\Delta$V} \\
\midrule
No dropout & 0.3623 & 0.3233 & 0.0826 & -27.20\% & 0.4497 & 0.4073 & 0.0527 & -8.29\% & 0.4883 & 0.4441 & 0.0453 & -- \\
\rowcolor{gray!15}
\textbf{w/ dropout} & \textbf{0.4792} & \textbf{0.4311} & \textbf{0.0456} & \textbf{-5.78\%} & \textbf{0.5027} & \textbf{0.4573} & \textbf{0.0437} & \textbf{-0.04\%} & \textbf{0.5030} & \textbf{0.4575} & \textbf{0.0436} & \textbf{--} \\
\bottomrule
\end{tabular}
}
\end{table}

\begin{figure}[t]
\centering
\includegraphics[width=\linewidth]{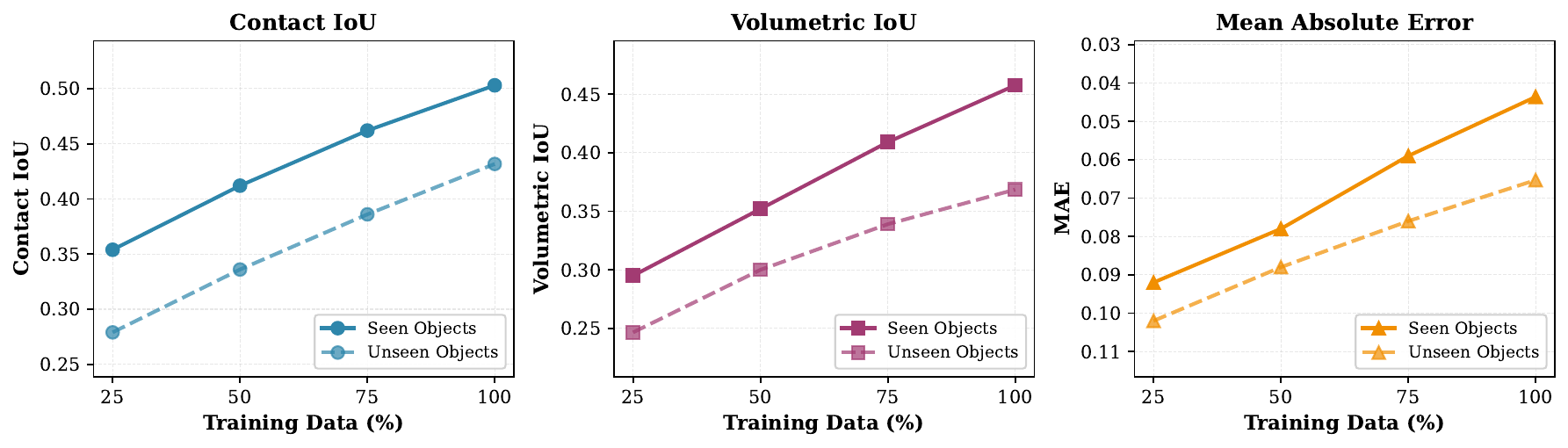}
\caption{Data scaling ablation study. Performance improves consistently with more training data across all metrics, demonstrating the model's ability to leverage larger datasets effectively.}
\label{fig:data_scaling}
\end{figure}

\textbf{Performance scales with data.}
We study how performance varies with the amount of training data by training the model on 25\%, 50\%, 75\%, and 100\% of the dataset. As shown in Figure~\ref{fig:data_scaling}, performance consistently improves as more data is used. In particular, both Contact IoU and Volumetric IoU exhibit steady gains, indicating that the model benefits from increased diversity in interaction patterns and contact configurations.
Notably, the improvement does not saturate at higher data regimes, suggesting that the proposed task remains \textbf{data-hungry and can further benefit from larger-scale tactile datasets}. This highlights the importance of scaling data collection for learning robust vision-to-tactile mappings.

\subsection{Qualitative Results}
\label{sec:qualitative}

\begin{figure}[!htb]
\centering
\includegraphics[width=0.9\textwidth]{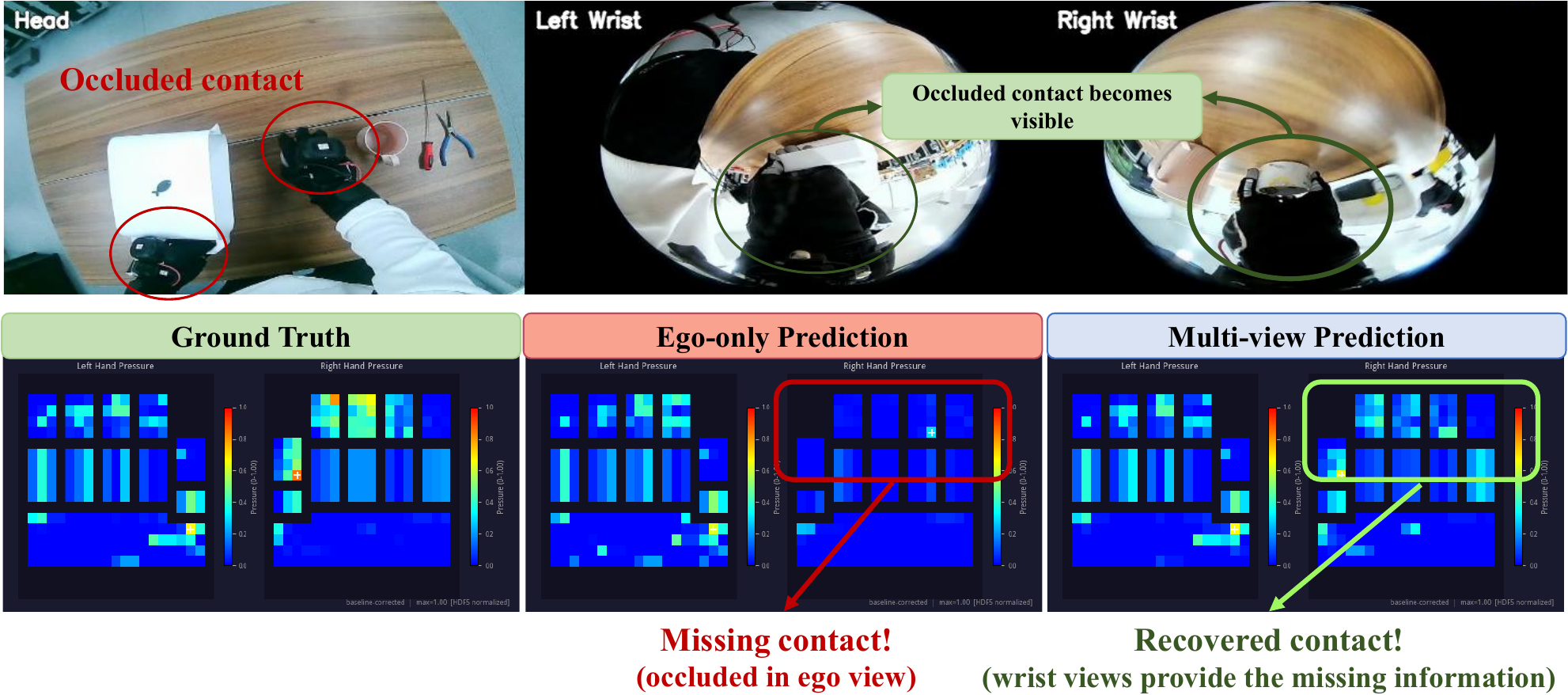}
\caption{\textbf{Multi-view wrist cameras recover occluded hand--object contact.}
Top: the egocentric view suffers from occlusion, while wrist-mounted views reveal the contact interface.
Bottom: ego-only prediction misses contact in occluded regions, whereas multi-view prediction recovers accurate pressure distributions consistent with the ground truth.}
\label{fig:qualitative}
\end{figure}

Figure~\ref{fig:qualitative} illustrates how occlusion in egocentric views leads to incomplete tactile predictions. When the contact interface is not directly visible, the model lacks sufficient visual evidence to infer pressure accurately, resulting in missing contact regions. By incorporating wrist-mounted views that directly observe the contact interface, the model gains access to previously occluded information and is able to recover both the location and intensity of contact. This demonstrates that multi-view observations provide critical complementary evidence for resolving occlusion-induced ambiguity in vision-to-tactile prediction.

\section{Conclusion}
\label{sec:conclusion}


We presented EgoTouch, a large-scale multi-view egocentric dataset with dense tactile supervision for bimanual hand-object interaction. EgoTouch provides synchronized head- and wrist-mounted RGB videos, bimanual 3D hand pose, tracker poses, and continuous tactile pressure maps across 208 tasks and 1,891 episodes. We further introduced TouchAnything, a multi-view vision-to-touch baseline with cross-view fusion and view dropout for flexible inference under different camera configurations. Experiments show that wrist-mounted views provide complementary contact evidence and improve tactile prediction, especially for contact localization and pressure estimation under egocentric occlusion. We hope EgoTouch will support future research on tactile-grounded embodied perception, manipulation, and learning from egocentric human interaction data.



\section{Limitations and Future Work}

Our work has several limitations. First, all current training data are collected with tactile gloves, which may introduce glove-specific appearance bias and limit generalization to bare-hand tactile estimation. Future work will explore glove-to-bare-hand retargeting and augmentation to improve robustness in natural human interactions.

Second, our data-scaling analysis shows that model performance has not yet saturated. As shown in Fig.~\ref{fig:data_scaling}, Contact IoU and Volumetric IoU continue to improve with more training data, suggesting that vision-to-touch prediction remains data-hungry. We therefore plan to expand EgoTouch with more diverse objects, environments, contact patterns, and manipulation behaviors.

Finally, the current benchmark focuses primarily on tactile estimation. We hope EgoTouch can support future research on tactile-grounded embodied intelligence, including contact-aware manipulation, grasp stability prediction, affordance learning, tactile-enhanced world models, and robot policy learning from egocentric human demonstrations.

\clearpage

\bibliographystyle{plainnat}
\bibliography{references}

\clearpage

\beginappendix
\section{Related Work}
\label{sec:related}
\subsection{Egocentric Hand-Object Interaction Datasets}

Recent egocentric datasets~\citep{DBLP:journals/access/SeoLLLGRK26,DBLP:journals/corr/abs-2603-13741,DBLP:journals/corr/abs-2603-25135,DBLP:conf/cvpr/PerrettDSEPPLGB25,DBLP:conf/cvpr/SuSMWY25} have substantially advanced the study of hand-object interaction from a first-person perspective. Ego4D~\citep{grauman2022ego4d} and EPIC-KITCHENS~\citep{damen2018epic} provide large-scale egocentric video for activity understanding, but do not offer the paired 3D hand pose and dense tactile annotations needed for tactile reasoning. EgoDex~\citep{hoque2026egodexlearningdexterousmanipulation} scales up egocentric manipulation data with 3D hand and finger tracking across 194 tasks, but relies on a single egocentric view and does not provide tactile annotations. EgoPressure~\citep{grady2024egopressure} pairs egocentric video with real pressure supervision, but focuses on single-hand hand-surface interactions collected in a controlled indoor setup, lacking diverse hand-object manipulation scenarios. OpenTouch~\citep{song2025opentouchbringingfullhandtouch} introduces in-the-wild full-hand tactile sensing with synchronized video-touch-pose data, but remains limited to single-hand interactions and a single first-person viewpoint.

A key limitation shared by these datasets is the lack of viewpoints that can directly observe hand-object contact regions, leading to severe occlusion of critical contact areas, especially the palmar surfaces where pressure is applied. While some datasets introduce additional views, they do not provide complementary perspectives that explicitly capture the contact interface. More importantly, no existing dataset jointly provides synchronized multi-view video, bimanual hand pose, and dense real tactile sensing. EgoTouch addresses this gap by combining a head-mounted egocentric camera with dual wrist-mounted cameras that directly observe contact regions, together with dense continuous pressure maps, enabling tactile prediction under realistic occlusion and viewpoint variation.

\subsection{Vision-to-Touch Prediction}

While hardware-based tactile sensors such as GelSight~\citep{yuan2017gelsight} and DIGIT~\citep{lambeta2020digit} provide high-resolution contact signals, they are difficult to deploy at scale. Vision-to-touch prediction has therefore emerged as an alternative, aiming to infer tactile feedback from visual observations. PressureVision~\citep{yang2022pressurevision} predicts hand pressure maps from a single RGB image using a convolutional network. VisGel~\citep{li2019visgel} learns cross-modal representations between vision and touch using paired GelSight data. Touching a NeRF~\citep{zhong2023touching} leverages neural radiance fields to synthesize tactile signals from 3D geometry. EgoPressure~\citep{grady2024egopressure} further explores pressure estimation in egocentric settings by predicting surface pressure from RGB observations, but focuses on relatively simple hand-surface interactions rather than diverse hand-object manipulation scenarios.

However, existing vision-to-touch approaches remain limited for realistic egocentric manipulation. In such settings, critical hand-object contact regions are frequently occluded, especially on the palmar surfaces. As a result, visual inputs often lack direct observations of the contact interface, regardless of model capacity. This limitation cannot be resolved by better models alone; it requires complementary viewpoints that explicitly capture the contact region. Building on EgoTouch, we propose TouchAnything, a multi-view tactile prediction model that leverages wrist-mounted cameras to directly observe contact regions, thereby enabling robust tactile estimation under occlusion and viewpoint variation.

\subsection{Multi-View Learning for Hand Interaction}

Multi-view learning~\citep{DBLP:journals/access/AryalGPSDC26,DBLP:conf/cvpr/BanerjeeSMHHZZF25} mitigates occlusion by leveraging complementary observations from different viewpoints. In hand understanding, it enables accurate 3D hand pose and mesh reconstruction by recovering geometry that is not visible from a single view~\citep{Hampali_2022_CVPR, DBLP:conf/iccv/ZimmermannCYRAB19, iqbal2018hand}. In egocentric perception, Ego-Exo4D~\citep{grauman2024egoexo4d} combines egocentric and exocentric views to improve activity understanding, highlighting the benefits of cross-view complementarity.

However, existing multi-view approaches primarily target geometric reconstruction or high-level action understanding, rather than modeling physical interaction signals such as tactile feedback. Moreover, while some datasets and methods incorporate multiple views, these viewpoints are typically external or global and do not directly observe the hand-object contact interface. As a result, critical contact regions—especially on the palmar surfaces—remain occluded or only indirectly inferred. In this work, we introduce wrist-mounted cameras that provide contact-aware viewpoints, directly capturing hand-object interactions from complementary perspectives. This design enhances tactile estimation by supplying visual evidence of contact regions that are otherwise difficult to infer from standard viewpoints.

\section{Additional Dataset Details}
\subsection{Data Collection Setup}
\label{sec:appendix_data_collection}
We collect a multimodal dataset of bimanual interactions using a wearable acquisition system that records synchronized RGB videos, tracker poses, bimanual hand kinematics, and dense tactile measurements at a target rate of 30Hz.

\paragraph{Hardware Configuration.}

The visual subsystem contains three RGB cameras: a chest/head-mounted egocentric camera and two wrist-mounted cameras observing the left and right hands. During acquisition, the GUI stores these views directly as three 30 FPS videos, \texttt{chest.mp4}, \texttt{left.mp4}, and \texttt{right.mp4}. The wrist cameras provide close-range observations of hand-object contacts that are often occluded in the egocentric view.

For global spatial localization, we use three HTC Vive Trackers assigned to the roles \texttt{chest}, \texttt{left\_wrist}, and \texttt{right\_wrist}. The trackers are read through OpenXR using the \texttt{XR\_HTCX\_vive\_tracker\_interaction} extension. For each collection tick, the latest valid tracker state is stored as a 3D translation and quaternion rotation in \texttt{vive\_poses.json}.

Bimanual hand pose is captured using Rokoko motion-capture gloves. The Rokoko stream is received through UDP, parsed into 21 3D joints per hand, and cached for low-latency access by the GUI. When calibration matrices are available, the system transforms the left and right Rokoko hand joints into the Vive coordinate frame using the corresponding wrist tracker pose and records this with an \texttt{aligned\_to\_vive} flag. The resulting per-frame hand joints are saved in \texttt{rokoko\_hands.json}.

Dense tactile feedback is captured by custom pressure-sensing gloves. Each hand sends a 256-channel 8-bit pressure vector through a serial connection at 921600 baud, together with an IMU packet decoded into a quaternion. The latest pressure vectors and IMU quaternions for both hands are saved in \texttt{jq\_pressure.json}.

\paragraph{Data Acquisition Pipeline.}

The current acquisition software is implemented in the PyQt GUI. Each recording episode is stored under a timestamped directory with the following structure:
\begin{verbatim}
<root>/<category>/<date>/<task>/<episode_timestamp>/
    chest.mp4
    left.mp4
    right.mp4
    jq_pressure.json
    rokoko_hands.json
    vive_poses.json
    camera_matrix.txt
\end{verbatim}
The three \texttt{.json} files are JSON Lines files: each line stores one frame-level record with a common timestamp \texttt{ts} and integer \texttt{frame\_index}. The saved records contain:
\begin{itemize}[leftmargin=*,itemsep=2pt]
    \item \texttt{jq\_pressure.json}: \texttt{sensor\_left} and \texttt{sensor\_right}, each a 256-value pressure vector, plus \texttt{quat\_left} and \texttt{quat\_right};
    \item \texttt{rokoko\_hands.json}: \texttt{left\_pos} and \texttt{right\_pos}, each a $21\times3$ joint array, plus \texttt{aligned\_to\_vive};
    \item \texttt{vive\_poses.json}: a dictionary of tracker poses for \texttt{chest}, \texttt{left\_wrist}, and \texttt{right\_wrist}, each containing \texttt{trans} and \texttt{rot}.
\end{itemize}
If camera intrinsics are available, they are written once as \texttt{camera\_matrix.txt}. This episode-level storage format avoids creating thousands of small image and text files and keeps the RGB streams temporally aligned with compact frame-wise sensor metadata.

\paragraph{Synchronization Strategy.}

The system uses software synchronization. Sensor readers for cameras, Rokoko, Vive Trackers, and pressure gloves run asynchronously and keep their latest valid measurements in memory. A 30Hz timer submits frame indices to a background saving worker. For each frame index, the GUI builds a snapshot containing the latest available RGB frames, hand joints, tracker poses, pressure vectors, and glove quaternions. The worker then appends one row to each JSON Lines file and writes the corresponding RGB frames to the video streams. When recording stops, the GUI waits for the queued frames to be written before closing the video and JSON files, so the number of saved JSON rows matches the number of submitted collection frames whenever possible.

\subsection{Tactile Grid Mapping and Preprocessing}
\label{sec:appendix_tactile_preprocessing}

The raw tactile glove stream contains 256 sensor values per hand. Although these values can be interpreted as a compact $16\times16$ array, the physical sensor layout on the glove is not a regular image grid: different sensors correspond to different fingers, palm regions, and bending/contact locations. To preserve the spatial structure of the hand, we remap the raw 256-dimensional tactile vector into a $21\times21$ hand-shaped pressure grid before training. The mapping is defined by hand-specific JSON files, where each key specifies a target grid coordinate $(r,c)$ and each value specifies the corresponding raw sensor index. This produces a sparse hand-shaped tactile map whose valid locations follow the physical arrangement of the glove sensors.

For each frame, we initialize a $21\times21$ grid with invalid locations marked as NaN and fill only the mapped sensor locations. The left hand is placed directly according to the mapping, while the right hand is horizontally mirrored so that left and right tactile maps share a consistent canonical hand coordinate system. This canonical representation makes the model output directly interpretable as a hand-shaped pressure distribution rather than an arbitrary sensor vector, providing stronger spatial priors for learning contact location and pressure magnitude.

We further apply several preprocessing steps to improve data quality. First, we optionally subtract the first-frame baseline pressure when the first frame is judged to be contact-free, either according to manual contact annotations or a low-pressure threshold fallback. This removes static sensor bias while avoiding over-correction when the sequence starts with an active contact. Second, known broken columns in the right-hand tactile grid are repaired by interpolation from neighboring valid columns. Third, tactile sensors and bending-related sensors are normalized separately, preventing high bending-sensor values from compressing the dynamic range of true contact-pressure sensors. The processed pressure grids, baseline-correction flags, grid size, and normalization metadata are saved into \texttt{pressure\_grids.npz} for downstream conversion and training.

\subsection{HDF5 Dataset Conversion}
\label{sec:appendix_hdf5_conversion}

After cleaning and tactile-grid preprocessing, we convert each trajectory into a compact HDF5 file for efficient training and inference. The conversion script expects each trajectory to contain synchronized videos from the three cameras, processed pressure grids, and hand-pose annotations. During conversion, RGB frames are decoded from \texttt{ego.mp4}, \texttt{left.mp4}, and \texttt{right.mp4}; when available, GPU-accelerated FFmpeg decoding is used, with OpenCV as a fallback. The final frame count is set to the minimum valid length across the three camera streams to keep all views temporally aligned.

Each HDF5 file stores a stable hierarchy of modalities. Metadata include the trajectory id, task name, number of frames, FPS, image resolution, and quality flags such as duplicated wrist-camera frames. The \texttt{images} group stores the three RGB streams. The \texttt{poses} group stores 7D Vive Tracker poses for the head/chest and both wrists. The \texttt{hands} group stores Rokoko hand joints when available, as well as WiLoR-based left/right 21-joint hand poses and validity masks. The \texttt{pressure} group stores the processed left and right $21\times21$ pressure grids together with preprocessing metadata. Optional glove masks are stored under a separate \texttt{masks} group when available.

For large-scale conversion, we use the batch conversion script with \texttt{64} trajectory workers, gzip compression level \texttt{4}, and \texttt{--skip\_existing} to avoid recomputing valid HDF5 files. The same script also supports regenerating trajectories listed in a bad-file list, filtered by failure reason, which is useful for repairing read errors or quality-control failures without reprocessing the entire dataset. This HDF5 format substantially reduces data-loading overhead and ensures that all experiments use the same cleaned, temporally aligned, and spatially remapped tactile representation.

\section{Additional Implementation Details}

\label{sec:appendix_method}
We propose a multi-view tactile prediction model that takes as input any subset of the three camera views and bimanual hand pose, and predicts dense bilateral pressure maps. The architecture is designed to gracefully handle missing views at inference time through a view dropout training strategy.

\subsection{Problem Formulation}
\label{sec:appendix_formulation}

Given a video clip of $T$ frames from any subset $\mathcal{V} \subseteq \{V^{ego}, V^{wL}, V^{wR}\}$ of available views and the corresponding bimanual hand pose sequence $\mathbf{P} \in \mathbb{R}^{T \times 42 \times 3}$, our goal is to predict the bilateral pressure maps $\hat{\mathbf{M}} \in \mathbb{R}^{T \times 2 \times 21 \times 21}$ for both hands at each timestep. The model must produce reasonable predictions regardless of which views are available.

\subsection{Multi-View Vision Encoder}
\label{sec:appendix_multiview_encoder}

\paragraph{Shared backbone with view embeddings.}
All views are processed by a shared DINOv2-ViT-B/14~\citep{oquab2023dinov2} backbone, which extracts $N = 256$ patch tokens of dimension $D = 768$ per frame. To enable the model to distinguish which camera a patch originates from, we add a learnable view embedding $\mathbf{e}_v \in \mathbb{R}^D$ for each view $v \in \{ego, wL, wR\}$:
\begin{equation}
\mathbf{F}_v = \text{DINOv2}(V_v) + \mathbf{e}_v, \quad \mathbf{F}_v \in \mathbb{R}^{T \times N \times D}
\end{equation}
Sharing the backbone across views reduces the parameter count from $3 \times 86\text{M}$ (separate encoders) to $86\text{M} + 3 \times 768$ (shared encoder + view embeddings), improving both efficiency and generalization.

\paragraph{Cross-view attention.}
Rather than performing expensive attention over all $N \times |\mathcal{V}|$ patch tokens, we extract a summary token for each view via global average pooling and apply a lightweight cross-view transformer~\citep{vaswani2017attention} over the $|\mathcal{V}|$ summary tokens:
\begin{equation}
\mathbf{s}_v = \text{MeanPool}(\mathbf{F}_v), \quad
[\hat{\mathbf{s}}_1, \ldots, \hat{\mathbf{s}}_{|\mathcal{V}|}] = \text{CrossViewTransformer}([\mathbf{s}_1, \ldots, \mathbf{s}_{|\mathcal{V}|}])
\end{equation}
This allows each view's summary to attend to summaries from other views, enabling complementary information exchange (e.g., the wrist view can inform the egocentric view about occluded contact regions).

\paragraph{Gated view fusion.}
The fused summary tokens are passed through a gating network that learns view-dependent importance weights:
\begin{equation}
w_v = \text{softmax}\big(\text{MLP}(\hat{\mathbf{s}}_v)\big), \quad
\mathbf{F}^{fused} = \sum_{v \in \mathcal{V}} w_v \cdot \mathbf{F}_v
\end{equation}
The output $\mathbf{F}^{fused} \in \mathbb{R}^{T \times N \times D}$ has the same shape as single-view features, ensuring compatibility with downstream modules.

\subsection{Temporal Modeling and Pose-Vision Fusion}
\label{sec:appendix_temporal_fusion}

\paragraph{Temporal transformer.}
A windowed temporal transformer~\citep{vaswani2017attention} is applied across the time dimension to capture manipulation dynamics:
\begin{equation}
\mathbf{H} = \text{TemporalTransformer}(\mathbf{F}^{fused}), \quad \mathbf{H} \in \mathbb{R}^{T \times N \times D}
\end{equation}

\paragraph{Pose encoder.}
The bimanual hand pose $\mathbf{P} \in \mathbb{R}^{T \times 42 \times 3}$ is encoded by a transformer-based pose encoder that produces per-joint features $\mathbf{G} \in \mathbb{R}^{T \times 42 \times D}$.

\paragraph{Pose-vision cross-attention fusion.}
Each joint token queries the visual patch tokens via cross-attention, enabling spatially grounded fusion:
\begin{equation}
\mathbf{Z} = \text{CrossAttn}(Q{=}\mathbf{G},\; K{=}\mathbf{H},\; V{=}\mathbf{H}), \quad \mathbf{Z} \in \mathbb{R}^{T \times 42 \times D}
\end{equation}
This allows each joint to attend to the visual patches most relevant to its spatial location and contact state.

\subsection{Joint-Level Tactile Decoder}
\label{sec:appendix_decoder}

The fused joint features $\mathbf{Z}$ are decoded into bilateral pressure maps. The 42-joint features are split into left-hand (joints 1--21) and right-hand (joints 22--42) groups, each decoded independently into a $21 \times 21$ pressure map via an MLP followed by a reshape operation:
\begin{equation}
\hat{\mathbf{M}}^{left}_t = \sigma\big(\text{MLP}(\mathbf{Z}^{left}_t)\big) \in [0, 1]^{21 \times 21}, \quad
\hat{\mathbf{M}}^{right}_t = \sigma\big(\text{MLP}(\mathbf{Z}^{right}_t)\big) \in [0, 1]^{21 \times 21}
\end{equation}
where $\sigma$ is the sigmoid function ensuring outputs are in $[0, 1]$.

\subsection{View Dropout Training Strategy}
\label{sec:appendix_view_dropout}

A critical design requirement is that the model must work with any available subset of views at inference time. To achieve this, we employ \emph{view dropout} during training: the egocentric view is always retained, while each wrist view is independently dropped with probability $p = 0.3$, matching the configuration used in our training script. This exposes the model to four possible input configurations:

\begin{itemize}[leftmargin=*,itemsep=1pt]
\item Ego only (both wrist views dropped)
\item Ego + left wrist
\item Ego + right wrist
\item All three views (no views dropped)
\end{itemize}

At inference time, the same model accepts whichever views are available without architectural changes or fine-tuning. This enables systematic evaluation under different camera configurations, including ego-only deployment and full multi-view inference.

\subsection{Training Objective and Optimization}
\label{sec:appendix_loss}

The training objective combines pixel-wise regression, sparsity-aware pressure fitting, and spatial regularization. Let $\hat{\mathbf{M}}$ and $\mathbf{M}$ denote the predicted and ground-truth tactile maps. We optimize
\begin{equation}
\mathcal{L} = \lambda_{mse}\mathcal{L}_{mse} + \lambda_{l1}\mathcal{L}_{l1} + \lambda_{tv}\mathcal{L}_{TV},
\end{equation}
where $\mathcal{L}_{mse}$ and $\mathcal{L}_{l1}$ measure pressure reconstruction error and $\mathcal{L}_{TV}$ encourages spatial smoothness in the predicted pressure maps. To reduce the tendency of the model to predict all-zero pressure maps under sparse contact supervision, pixels with pressure greater than $0.1$ are treated as contact regions and assigned a larger loss weight. In our experiments, we use $\lambda_{mse}=1.0$, $\lambda_{l1}=0.5$, $\lambda_{tv}=0.01$, and a contact-region weight of $3.0$.

We train the model with AdamW using a learning rate of $5\times10^{-5}$, weight decay of $0.05$, and betas $(0.9, 0.999)$. The learning rate follows a cosine schedule with 10 warmup epochs and a minimum learning rate of $10^{-6}$. Training runs for 25 epochs. We use distributed data parallel training launched by \texttt{torchrun}; by default, the launcher uses six GPUs with per-GPU batch size 16 and gradient accumulation over 3 steps, resulting in an effective batch size of 288. The DINOv2 ViT-B/14 visual encoder is initialized from pretrained weights and kept frozen during training. We use clips of 8 frames sampled every 2 frames, RGB inputs resized to $224\times224$, 42 bimanual hand joints from WiLoR, glove color augmentation with probability 0.2, and $21\times21$ tactile maps aligned with the native pressure-sensor grid.

\subsection{Inference and Evaluation Protocol}
\label{sec:appendix_inference_protocol}

For evaluation, we load the best model checkpoint and run batched inference with the same configuration file used during training. The inference script supports both full evaluation and a lightweight mode for quick inspection. In the full setting, it evaluates up to 1000 trajectories per split; in lightweight mode, it samples one trajectory per task category from the split file. Unless otherwise specified, inference uses batch size 64, two worker processes, 30 FPS visualization, and skips saving HDF5 outputs to reduce storage usage while retaining videos and metrics.

The inference pipeline supports configurable view subsets through a \texttt{views} argument. We use this to evaluate ego-only input, single-wrist variants, and full multi-view input with the same trained model. Outputs are organized by configuration name, checkpoint name, view setting, and dataset split. For each evaluated split, the script saves visualizations and computes tactile prediction metrics including Temporal Accuracy, Contact IoU, Volumetric IoU, and MAE. This shared inference protocol ensures that all view configurations are compared under the same checkpoint, dataset split, preprocessing, and metric implementation.

\section{Additional Qualitative Results}

We provide additional qualitative examples of tactile prediction results across diverse manipulation tasks. Each visualization shows a 2×3 grid of frames sampled uniformly from the video sequence, with the egocentric RGB input (top row), ground truth pressure maps for both hands (middle row), and predicted pressure maps (bottom row). The model accurately captures contact locations, pressure intensity, and temporal dynamics across a wide range of interactions.

\begingroup
\captionsetup{font=footnotesize,skip=2pt}
\captionsetup[subfigure]{font=footnotesize,skip=1pt}

\begin{figure}[p]
\centering
\begin{subfigure}[t]{\linewidth}
\centering
\includegraphics[width=0.92\linewidth]{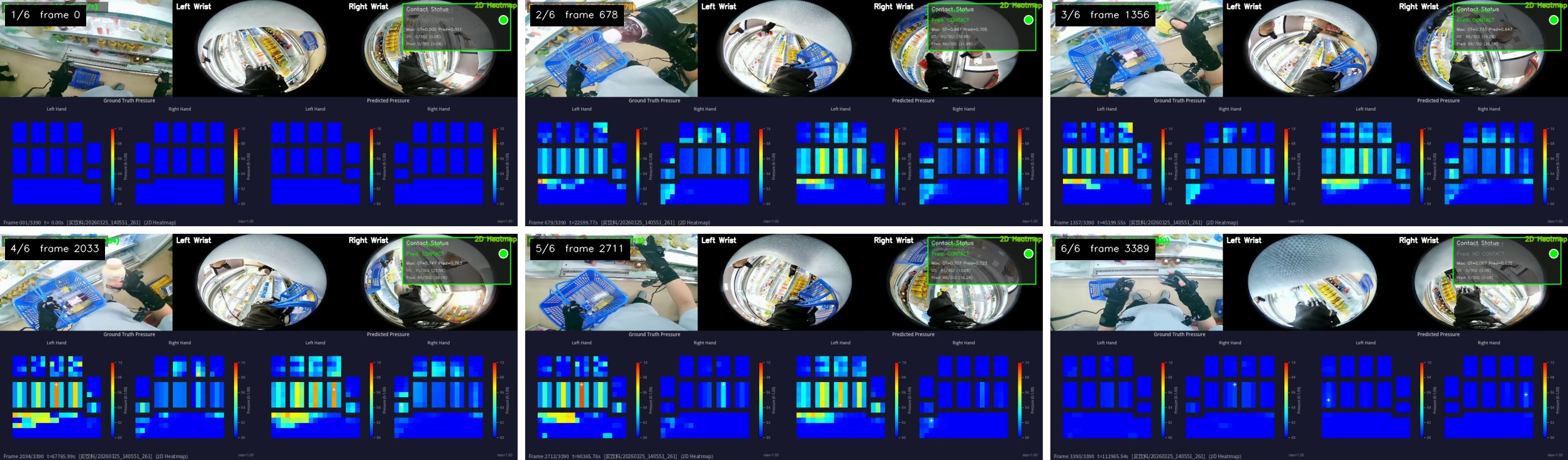}
\caption{Purchasing a beverage}
\end{subfigure}
\vspace{0.08cm}

\begin{subfigure}[t]{\linewidth}
\centering
\includegraphics[width=0.92\linewidth]{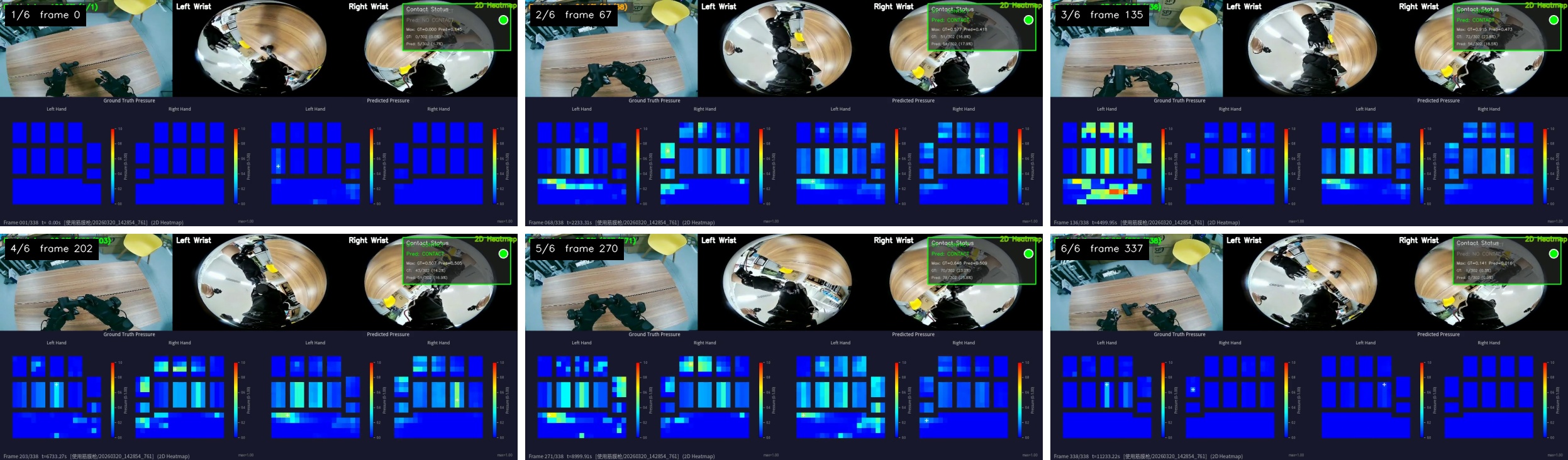}
\caption{Using a massage gun}
\end{subfigure}
\vspace{0.08cm}

\begin{subfigure}[t]{\linewidth}
\centering
\includegraphics[width=0.92\linewidth]{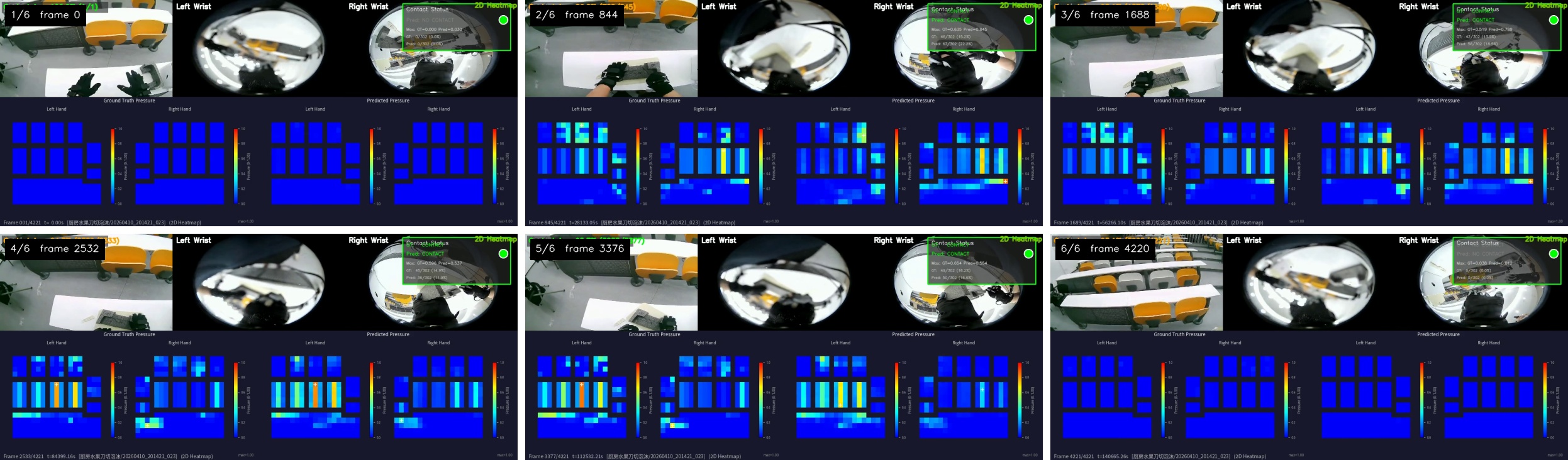}
\caption{Cutting foam with a kitchen knife}
\end{subfigure}
\vspace{0.08cm}

\begin{subfigure}[t]{\linewidth}
\centering
\includegraphics[width=0.92\linewidth]{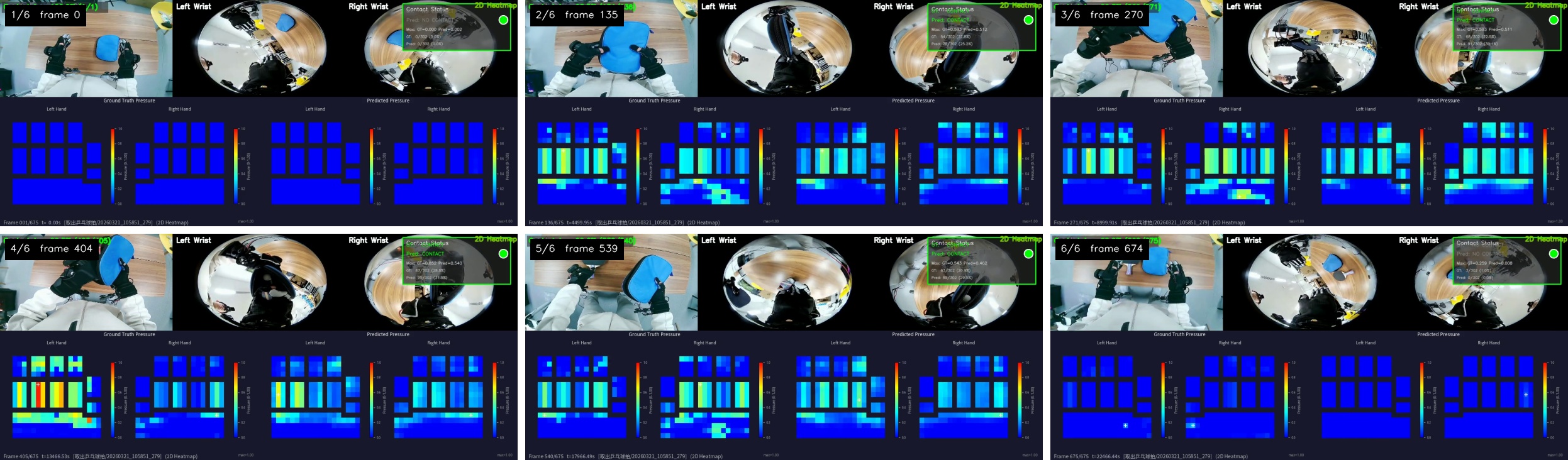}
\caption{Retrieving a ping-pong paddle}
\end{subfigure}

\caption{Tactile prediction results (1--4). Each subfigure shows a 2×3 grid of frames with predicted tactile pressure maps overlaid on the egocentric RGB input.}
\label{fig:qualitative_results_1}
\end{figure}

\begin{figure}[p]
\centering
\begin{subfigure}[t]{\linewidth}
\centering
\includegraphics[width=0.92\linewidth]{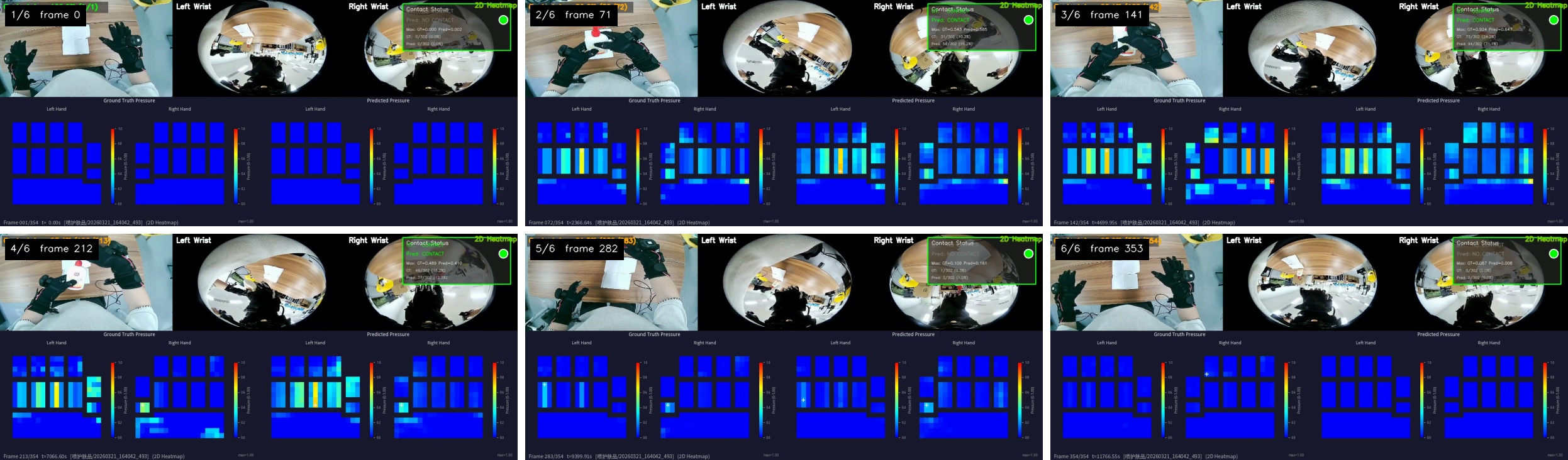}
\caption{Spraying skincare product}
\end{subfigure}
\vspace{0.08cm}

\begin{subfigure}[t]{\linewidth}
\centering
\includegraphics[width=0.92\linewidth]{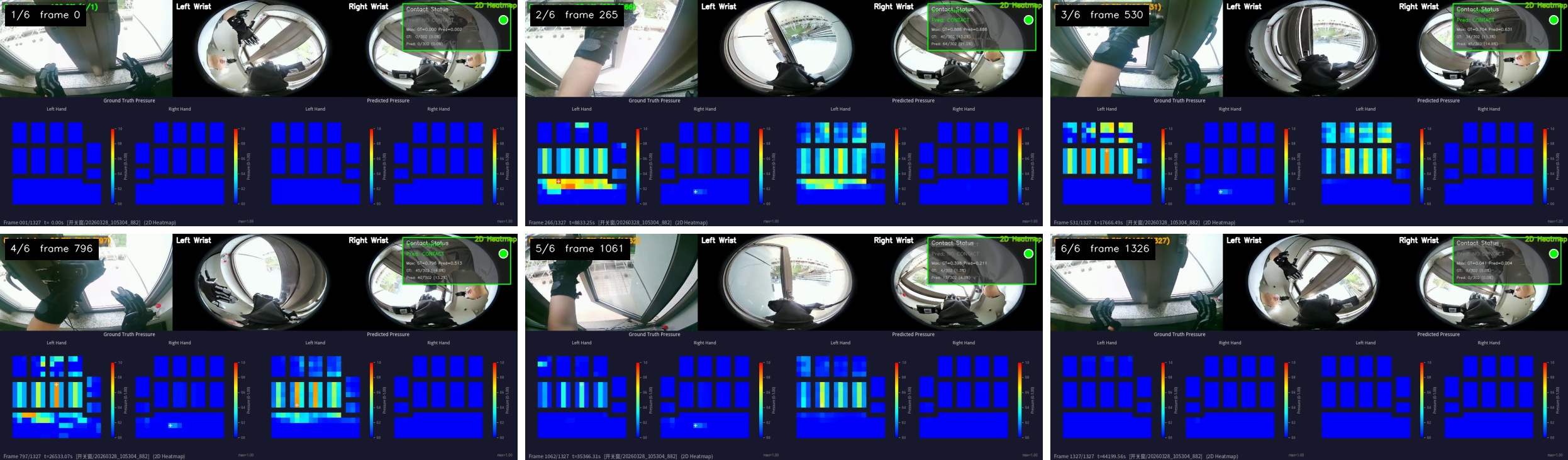}
\caption{Opening/closing a window}
\end{subfigure}
\vspace{0.08cm}

\begin{subfigure}[t]{\linewidth}
\centering
\includegraphics[width=0.92\linewidth]{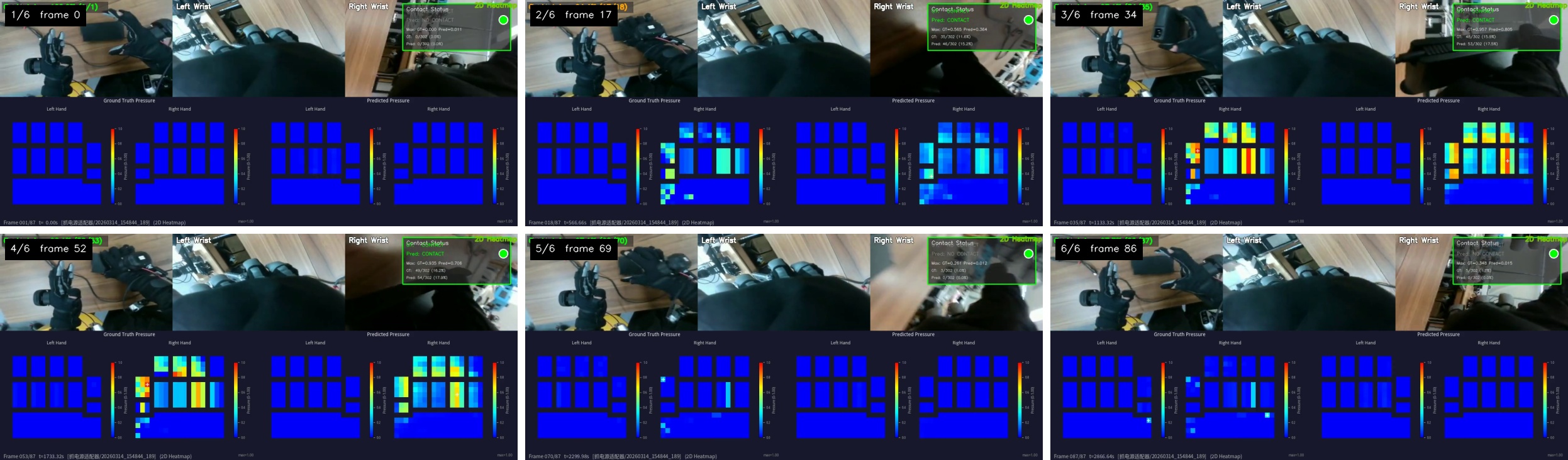}
\caption{Grasping a power adapter}
\end{subfigure}
\vspace{0.08cm}

\begin{subfigure}[t]{\linewidth}
\centering
\includegraphics[width=0.92\linewidth]{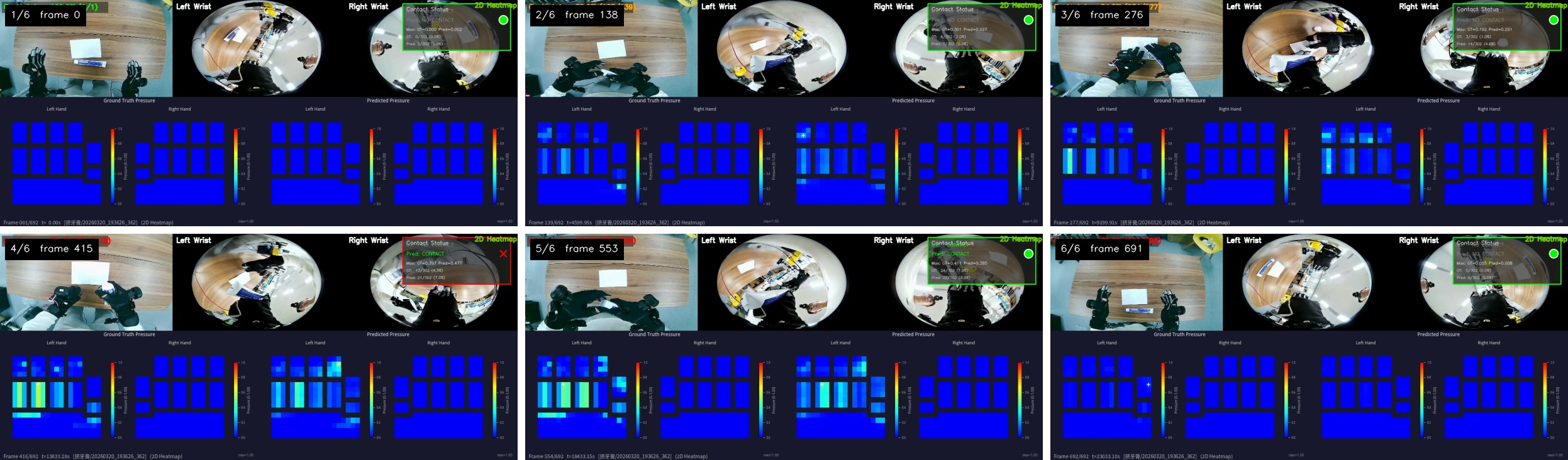}
\caption{Squeezing toothpaste}
\end{subfigure}

\caption{Tactile prediction results (5--8). The model captures fine-grained contact patterns and bimanual coordination across diverse manipulation tasks.}
\label{fig:qualitative_results_2}
\end{figure}

\begin{figure}[p]
\centering
\begin{subfigure}[t]{\linewidth}
\centering
\includegraphics[width=0.92\linewidth]{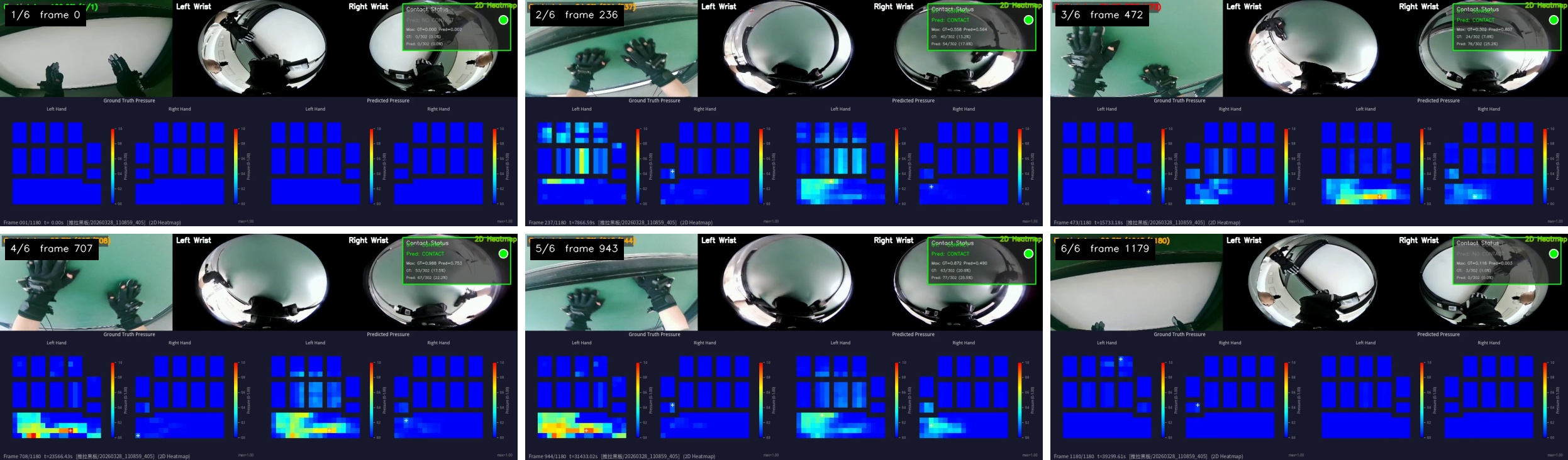}
\caption{Pushing/pulling a blackboard}
\end{subfigure}
\vspace{0.08cm}

\begin{subfigure}[t]{\linewidth}
\centering
\includegraphics[width=0.92\linewidth]{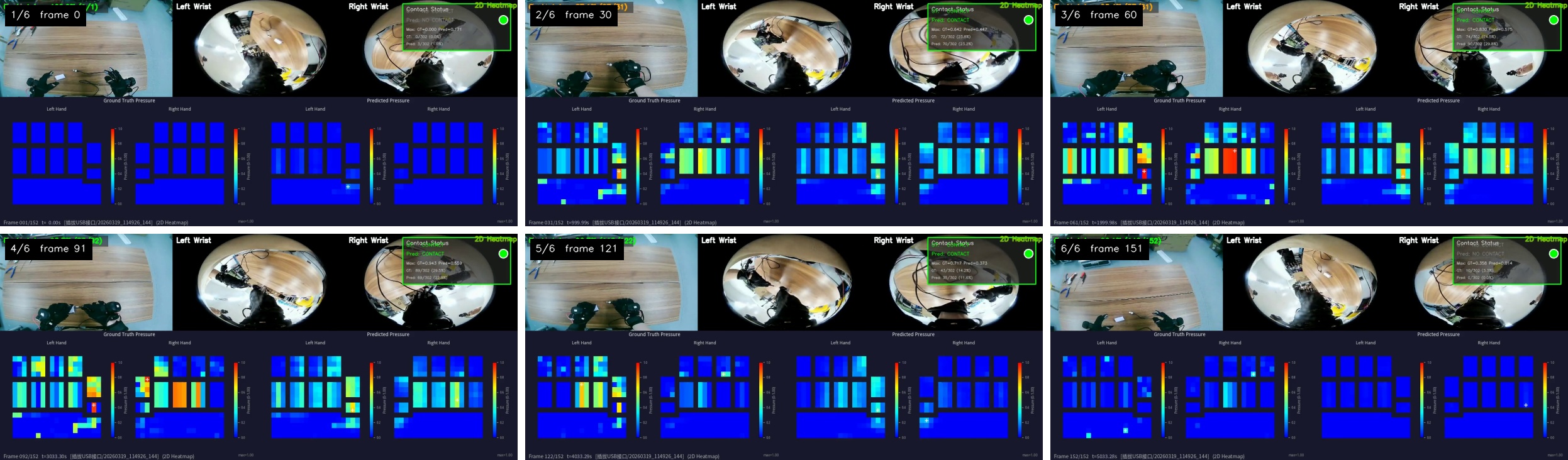}
\caption{Plugging/unplugging a USB connector}
\end{subfigure}
\vspace{0.08cm}

\begin{subfigure}[t]{\linewidth}
\centering
\includegraphics[width=0.92\linewidth]{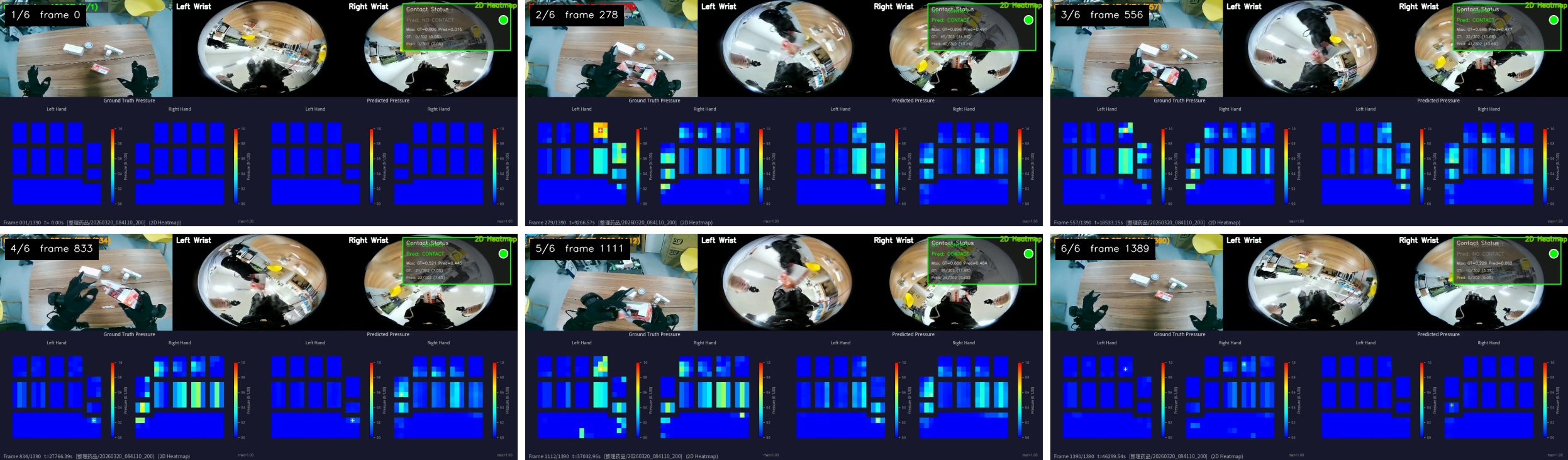}
\caption{Organizing medicine bottles}
\end{subfigure}
\vspace{0.08cm}

\begin{subfigure}[t]{\linewidth}
\centering
\includegraphics[width=0.92\linewidth]{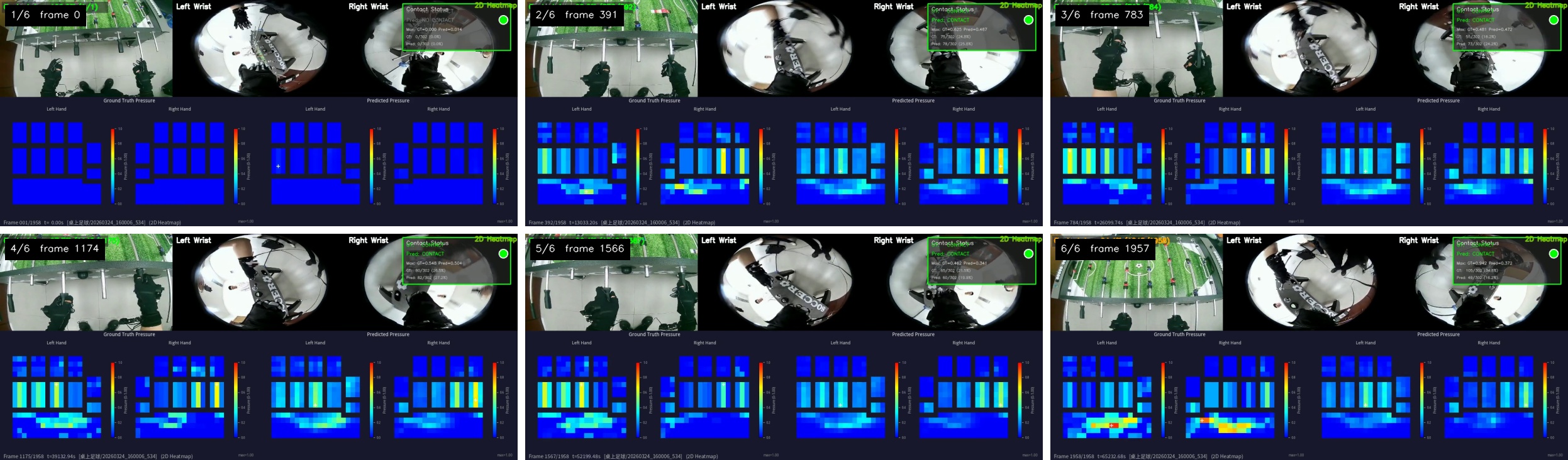}
\caption{Playing table football}
\end{subfigure}

\caption{Tactile prediction results (9--12). The model accurately predicts contact locations during precision manipulation and dynamic gameplay.}
\label{fig:qualitative_results_3}
\end{figure}

\begin{figure}[p]
\centering
\begin{subfigure}[t]{\linewidth}
\centering
\includegraphics[width=0.92\linewidth]{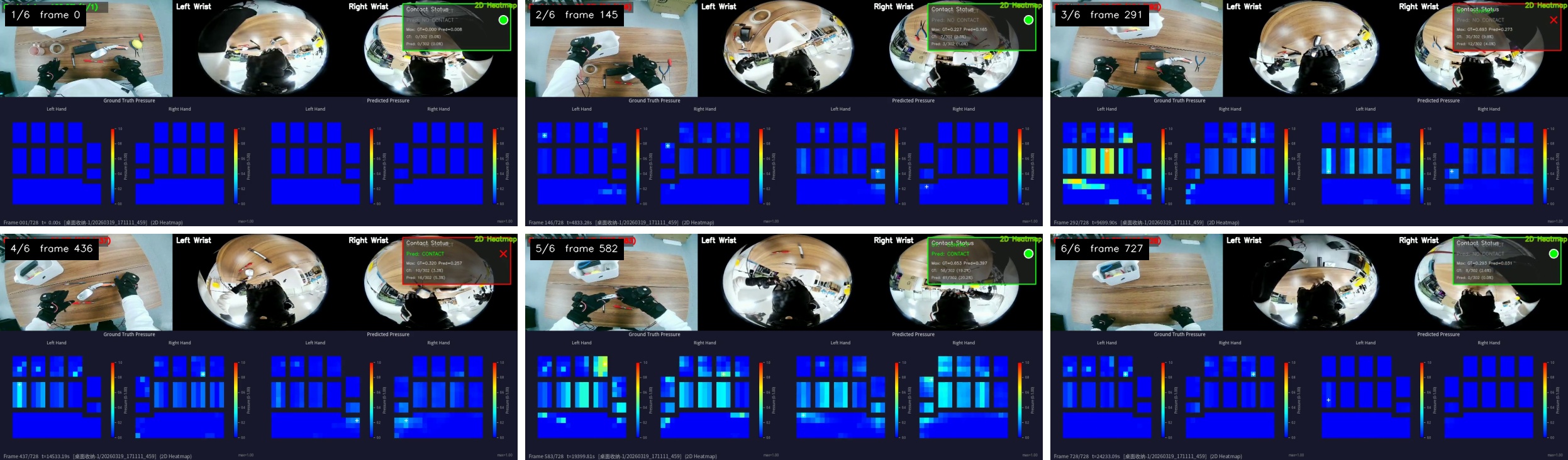}
\caption{Desktop organization}
\end{subfigure}
\vspace{0.08cm}

\begin{subfigure}[t]{\linewidth}
\centering
\includegraphics[width=0.92\linewidth]{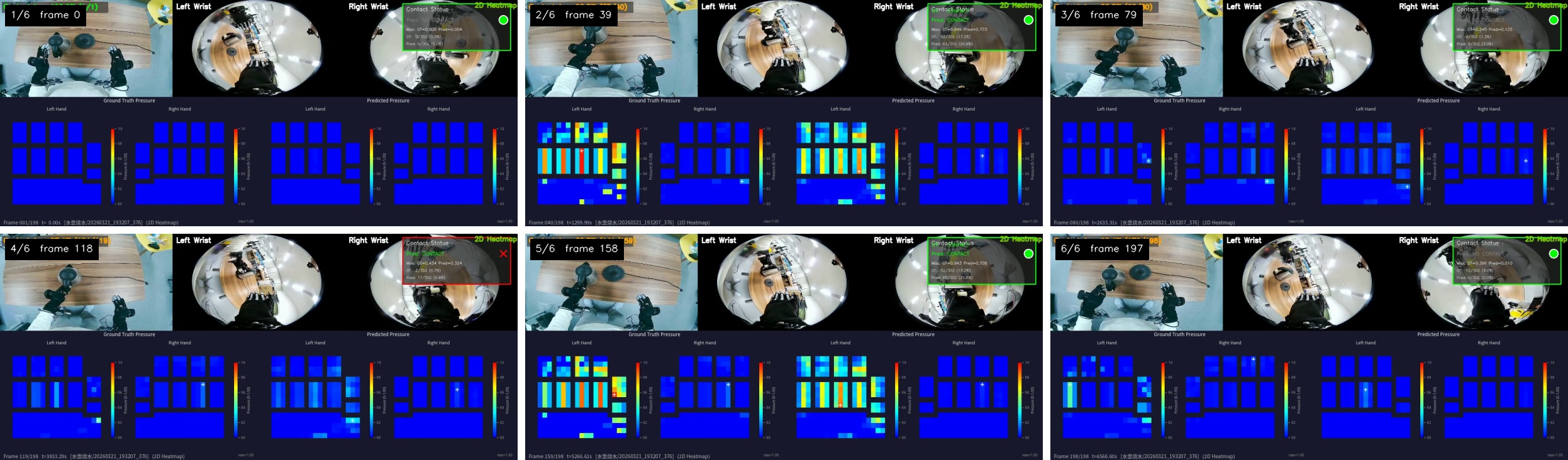}
\caption{Boiling water with a kettle}
\end{subfigure}
\vspace{0.08cm}

\begin{subfigure}[t]{\linewidth}
\centering
\includegraphics[width=0.92\linewidth]{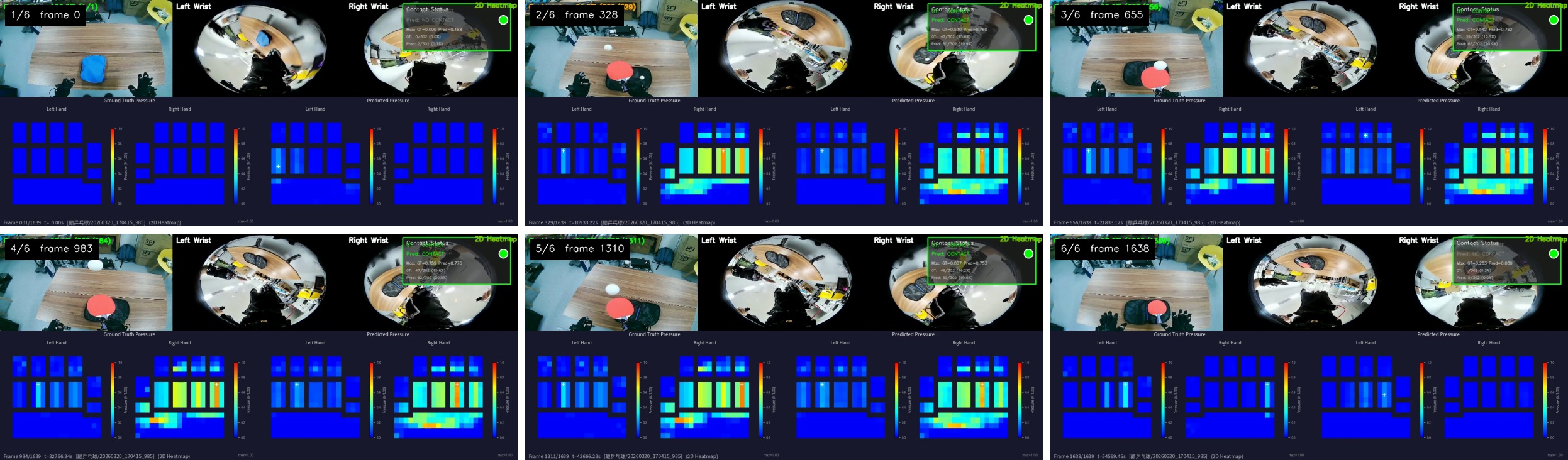}
\caption{Bouncing a ping-pong ball}
\end{subfigure}
\vspace{0.08cm}

\begin{subfigure}[t]{\linewidth}
\centering
\includegraphics[width=0.92\linewidth]{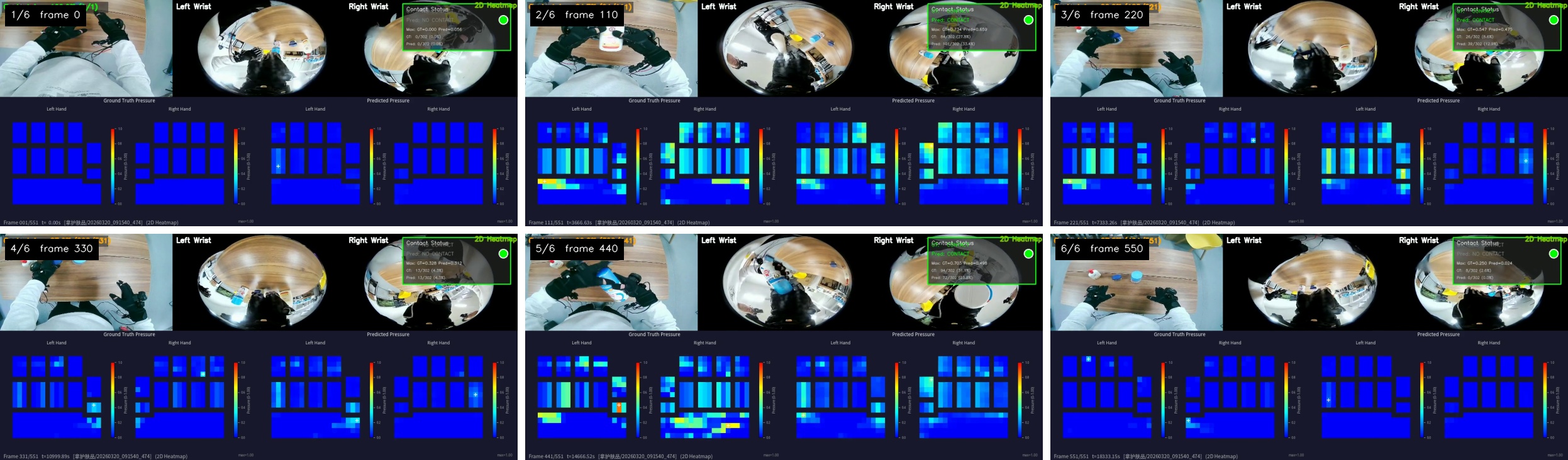}
\caption{Grasping skincare products}
\end{subfigure}

\caption{Tactile prediction results (13--16). The predicted pressure maps maintain spatial consistency with hand pose and visual observations while capturing temporal evolution of contact.}
\label{fig:qualitative_results_4}
\end{figure}

\endgroup

These examples demonstrate the model's ability to generalize across diverse object categories, manipulation strategies, and contact configurations. The predicted pressure maps maintain spatial consistency with the hand pose and visual observations, while capturing the temporal evolution of contact during continuous manipulation.

\subsection{Failure Cases and Limitations}

\begin{figure}[h]
\centering
\includegraphics[width=\linewidth]{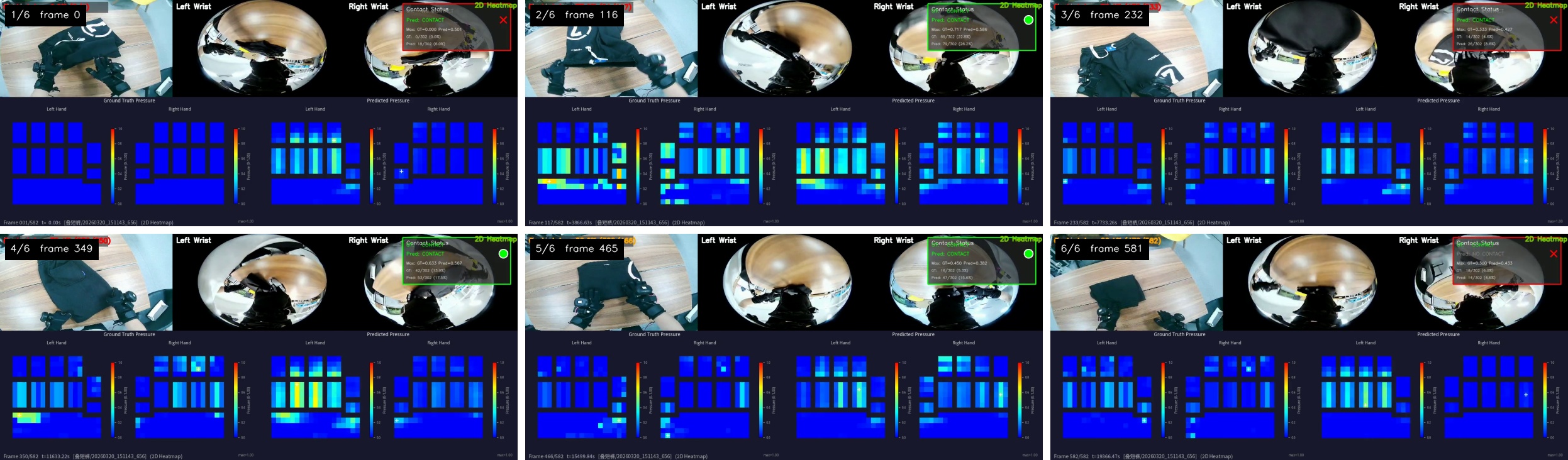}
\caption{Failure case: folding black shorts. In the first frame, the model incorrectly predicts contact on the left hand even though no contact has occurred. The black glove and black shorts create a low-contrast visual appearance that resembles contact, causing the model to hallucinate pressure. This highlights a key limitation: the model relies heavily on visual cues and can be confused by color similarity between the hand and object.}
\label{fig:appendix_failure_black}
\end{figure}

While the model achieves strong performance on most manipulation tasks, we observe failure modes in challenging visual conditions. Figure~\ref{fig:appendix_failure_black} shows a representative failure case where the model incorrectly predicts contact in the first frame when no contact has occurred yet.

This failure is caused by the low visual contrast between the black glove and the black shorts. From the egocentric view, the hand appears to be in close proximity to or overlapping with the dark fabric, creating an ambiguous visual signal that the model interprets as contact. This demonstrates that the model has learned to associate visual proximity and occlusion patterns with tactile contact, but can be misled when color similarity makes it difficult to distinguish hand-object boundaries.

Such failures suggest several directions for improvement: (1) incorporating explicit depth or hand-object segmentation to disambiguate proximity from contact, (2) augmenting training data with more challenging color combinations, and (3) leveraging temporal consistency to suppress isolated false-positive predictions in the first frame when no prior contact history exists.

\end{CJK*}
\end{document}